\documentclass{article}
\PassOptionsToPackage{numbers,compress}{natbib}
\usepackage[preprint]{neurips_2023}

\newcommand{\bn}[1]{\textsf{\textbf{#1}}}
\newcommand{\NA}{\multicolumn{1}{c}{---}}

\usepackage[utf8]{inputenc}
\usepackage[T1]{fontenc}
\usepackage{hyperref}
\usepackage{url}
\usepackage{booktabs}
\usepackage{amsfonts}
\usepackage{nicefrac}
\usepackage{microtype}
\usepackage{xcolor}

\usepackage{graphicx}
\usepackage{multirow}
\usepackage{multicol}
\usepackage{subcaption}
\usepackage{amsmath}

\title{Exploring the Robustness of Large Language Models for Solving Programming Problems}

\author{%
\bf Atsushi Shirafuji$^\dag$\thanks{Corresponding: m5261161@u-aizu.ac.jp} , Yutaka Watanobe$^{\dag}$, Takumi Ito$^\ddag$, Makoto Morishita$^\S$, \\
\bf Yuki Nakamura$^\ddag$, Yusuke Oda$^\ddag$, Jun Suzuki$^\ddag$ \\
\\
University of Aizu$^\dag$\\
Tohoku University$^\ddag$\\
NTT Communication Science Laboratories$^\S$
}

\begin{document}

\maketitle

\begin{abstract}
Using large language models (LLMs) for source code has recently gained attention. LLMs, such as Transformer-based models like Codex and ChatGPT, have been shown to be highly capable of solving a wide range of programming problems.
However, the extent to which LLMs understand problem descriptions and generate programs accordingly or just retrieve source code from the most relevant problem in training data based on superficial cues has not been discovered yet.
To explore this research question, we conduct experiments to understand the robustness of several popular LLMs, CodeGen and GPT-3.5 series models, capable of tackling code generation tasks in introductory programming problems.
Our experimental results show that CodeGen and Codex are sensitive to the superficial modifications of problem descriptions and significantly impact code generation performance.
Furthermore, we observe that Codex relies on variable names, as randomized variables decrease the solved rate significantly.
However, the state-of-the-art (SOTA) models, such as InstructGPT and ChatGPT, show higher robustness to superficial modifications and have an outstanding capability for solving programming problems.
This highlights the fact that slight modifications to the prompts given to the LLMs can greatly affect code generation performance, and careful formatting of prompts is essential for high-quality code generation, while the SOTA models are becoming more robust to perturbations.
\footnote{This paper is an extended work of an earlier version presented at SOMET 2022~\citep{shirafuji2022prompt}.}
\end{abstract}

\section{Introduction}
\label{sec:introduction}

Recent advances in natural language processing (NLP) have led to the development of large language models
(LLMs or large LMs\footnote{In this paper, we basically use the abbreviation LLM because we cover large-scale language models. However, when discussing non-large-scale language models, such as when discussing prior research, the abbreviation LM is used instead.}).
Such LLMs have often been trained on datasets containing the source code (programs), and thus they can also generate human-like source code.
These models can be applied in various code-related tasks to assist developers, such as
generating source code from natural language (code generation)~\citep{chen2021codex, parvez2021retrieval, nijkamp2022codegen, fried2022incoder, li2022alphacode, chen2022codet, chowdhery2022palm, christopoulou2022pangucoder, le2022coderl, xu2022polycoder, wang2021codet5, wei2019dual},
generating natural language from source code (code summarization)~\citep{chen2021codex, parvez2021retrieval, fried2022incoder, wang2021codet5, wei2019dual, lu2021codexglue},
translating source code from one programming language into another (code translation)~\citep{lu2021codexglue, ahmad2021plbart, roziere2020transcoder, roziere2022transcoder-st},
generating source code to complete a partially written program (code completion)~\citep{chen2021codex, nijkamp2022codegen, fried2022incoder, li2022alphacode, lu2021codexglue, svyatkovskiy2020intellicode, terada2021completion},
detecting and correcting errors in source code (e.g., vulnerability detection and program repair)~\citep{lu2021codexglue, berabi2021tfix, rahman2021repair, matsumoto2021repair, prenner2021repair, pearce2021securitybugs, joshi2022repair},
and classifying approaches or algorithms used in source code (e.g., program classification)~\citep{shalaby2017algorithm, chourasia2022algorithm, watanobe2023algorithm}.

However, the performance of these models depends on the quality of the prompts they receive, as they are originally trained to predict the next word or token given a sequence of previous ones.
In practice, LLMs' users need to carefully construct prompts to effectively leverage the LLMs; the phase investigating the prompts to obtain better generation results is called \textit{prompt engineering}.
As the job title of a \textit{prompt engineer} has been opened recently with a high range of salary, designing a better prompt template is considered essential for better generation in current LLMs.
Therefore, a comprehensive grasp of the robustness of these models to the given prompts is crucial to effectively leverage their capabilities.

The development in this field is extremely fast, and the interpretations of LLMs are not straightforward, which means that the specific accomplishments of current state-of-the-art (SOTA) LLMs are not thoroughly explained.
More specifically, it has not been sufficiently clarified whether LLMs understand queries and source code they receive and generate.
One possible explanation for their exceptional performance could be that, because of the high capacity of LLMs to memorize training data, they can perfectly respond to queries when identical queries appear in the training data and are being memorized by the LLMs but not otherwise.

In the context of natural language generation, memorization of training data can be a critical issue regarding leaking personal information~\citep{huang2022leaking}.
In addition, in the context of code generation, copying some amount of source code from open-source projects can lead to the open-source license infringement issue~\citep{ciniselli2022copy}.
Several works have investigated the copying and memorization issues of the texts generated by GPT-based models~\citep{carlini2021extract,elangovan2021memorization}.
As related work, \citet{karmakar2022hackerrank} demonstrated the potential of memorization issues of Codex against programming problems from HackerRank\footnote{\url{https://www.hackerrank.com/}.}; \citet{mastropaolo2023robustness} investigated code generation using different but semantically equivalent natural language descriptions; and \citet{wang2022recode} proposed ReCode, a benchmark for evaluating the robustness of code generation LMs.
Although several prior works investigated the capability and limitation of current GPT-based models from the perspectives of memorization, copying, and robustness, to the best of our knowledge, there has been no systematic study on the impact of prompts on the performance of LLMs in solving programming problems.

To remove doubts and gain a better understanding of the current progress of SOTA LLMs, we explore their ability to generate Python programs.
We focus on the task of automatically solving programming problems.
As a case study, four popular recent SOTA code generation LLMs, CodeGen~\citep{nijkamp2022codegen}, Codex~\citep{chen2021codex}, InstructGPT~\citep{ouyang2022instructgpt}, and ChatGPT\footnote{\url{https://openai.com/blog/chatgpt/}.}, are selected as automatic programming problem solvers.
From various perspectives, we assess the LLMs' capabilities and limitations in understanding problem descriptions and the corresponding generated programs.

We primarily perform two experiments: (1) formatting problem descriptions and (2) modifying problem descriptions. 
In the first experiment, we format problem descriptions using several rules.
We refer to this as \textit{superficial modification} since formatting does not alter the problem specification.
In contrast, the second experiment alters problem descriptions, which may change the difficulty or solution approach, including both \textit{superficial and semantic modifications}.
The investigation is based on comparing the generated programs by partially modifying the prompts given to the LLMs.

For experiments, we use the problem descriptions of introductory programming problems provided on Aizu Online Judge (AOJ)~\citep{watanobe2004aoj, watanobe2022aoj}.
We also use hidden test cases to validate the functional correctness of the generated programs.

The experimental results show that the earlier models, CodeGen and Codex, may not understand the queries and source code in some cases, because the performance is influenced by the formatting and modifications of problem descriptions, even if the models perfectly handle some of the programming problems.
On the other hand, the latest models, InstructGPT and ChatGPT, which incorporate reinforcement learning from human feedback (RLHF)~\citep{ouyang2022instructgpt}, show almost the same performance for several modifications in problem descriptions, exhibiting robustness to the modifications.

The main contributions of our work are to fill the gap and provide insights for developers and researchers on how LLMs are influenced by various prompts (perturbations) for solving programming problems, as well as warning about the pitfalls of using LLMs in practice.

The rest of this paper is organized as follows.
Section~\ref{sec:solving-programming-problems} introduces the target task of this work, solving programming problems.
Section~\ref{sec:research-question} defines the primary research question (RQ) that we investigate throughout this work.
Section~\ref{sec:experiments} describes the methodology of how we organize the work to answer the RQ.
Section~\ref{sec:results} presents and discusses the obtained results.
Section~\ref{sec:related-work} reviews related work and provides an overview of LMs for code generation.
Finally, Section~\ref{sec:conclusion} concludes this paper.

The original problem descriptions can be accessed via the API\footnote{\url{http://developers.u-aizu.ac.jp/}.}.
The problem descriptions used in the experiments are available here\footnote{\url{https://github.com/ashirafj/aoj-formatted-problems}.}.

\section{Solving Programming Problems}
\label{sec:solving-programming-problems}

\begin{figure*}[t]
    \centering
    \includegraphics[width=\linewidth]{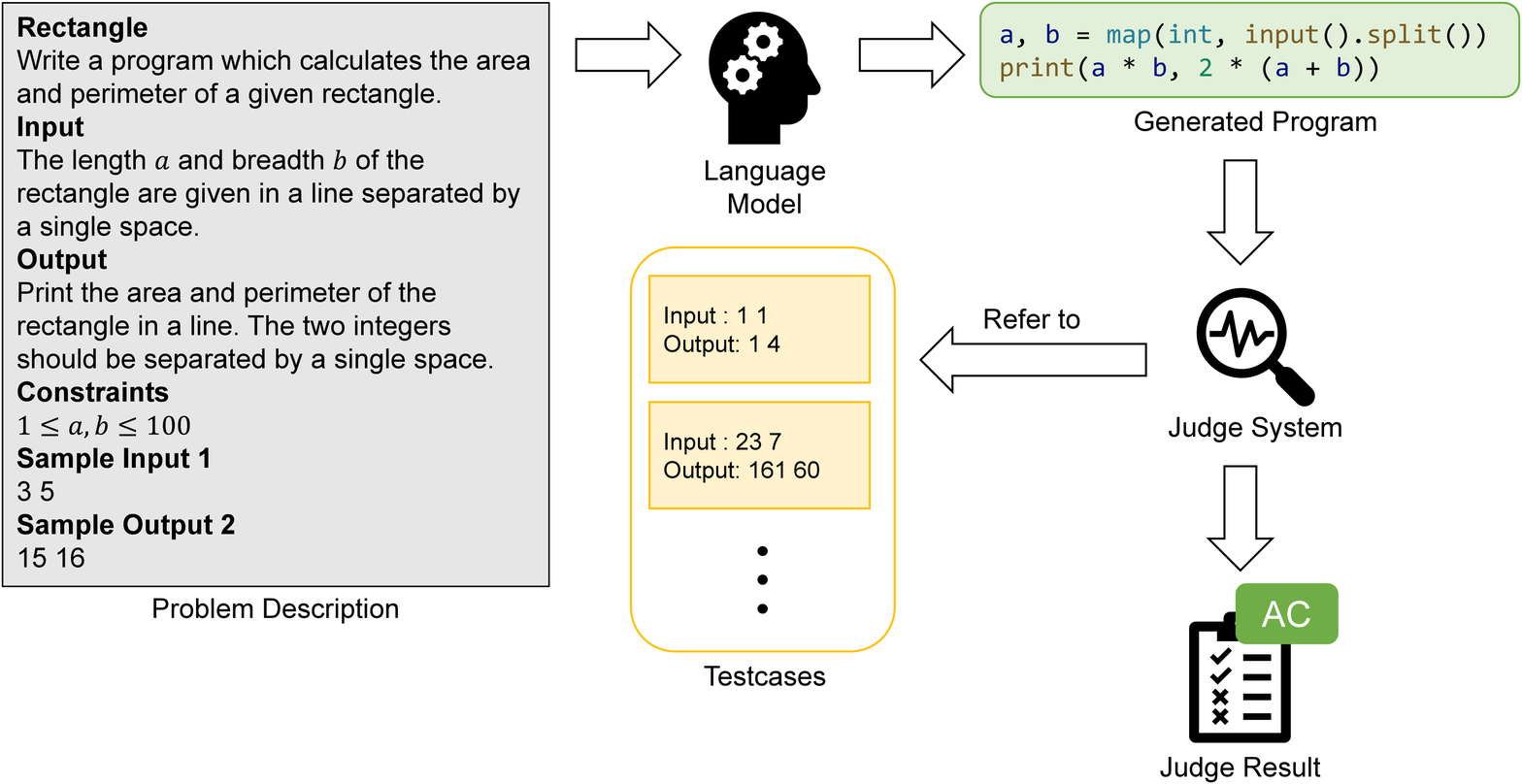}
    \caption{Task of automatically solving programming problems by an LM evaluated by a judge system based on hidden test cases, an example of a problem named ITP1\_1\_C from AOJ.}
    \label{fig:solving-programming-problems}
\end{figure*}

The task of solving programming problems is to generate complete solution programs from scratch in a specific programming language to solve a given problem using a problem description (Figure~\ref{fig:solving-programming-problems}).
Programming problems vary in type and difficulty, from introductory problems, such as input/output processing, simple calculations, if-branches, and for-loops, to advanced problems, such as algorithms and data structures requiring computer science knowledge.

The correctness of the submitted program is usually evaluated by actually executing the program using hidden test cases on judge systems~\citep{wasik2018oj}, similar to unit testing in software development.
The program is judged correct if it passes all hidden test cases and incorrect otherwise (i.e., if one or more test cases fail). 
The difference from unit testing is that not only the correctness of the output result of the program but also its execution time and memory usage are considered in the evaluation.
Each problem imposes different limitations according to the algorithms and constraints, and the limitations are loosened in each programming language for fairness.
For example, the time limit is 4~seconds and the memory limit is 655~MB when using Python~3, but 1~second and 131~MB when using C++ in ITP1\_1\_A on AOJ%
\footnote{In ITP1\_1\_A, the time limit is set to 1~second and the memory limit to 131,072~KB by default\footnotemark[9]. When using Python~3, these limits are loosened by $\times 4.0$ for time and $\times 5.0$ for memory, whereas neither of them is loosened in C++\footnotemark[10].}%
\footnote{\url{https://onlinejudge.u-aizu.ac.jp/courses/lesson/2/ITP1/1/ITP1_1_A}}%
\footnote{\url{https://onlinejudge.u-aizu.ac.jp/system_info}}%
.

Solvers must consider the constraints to select the suitable algorithm and data structure to satisfy the execution time and memory usage. 
A problem description contains the necessary information to solve the problem: a problem statement, input and output formats, variable constraints (e.g., value types and ranges), and example test cases.
However, since the problems for this task are usually based on competitive programming contests, problem descriptions rarely contain hints or instructions that clearly and directly explain how to solve the problem.
Solvers are required to decide on the solution approach to the given problem and then implement it as an executable program.

The results (i.e., the output of the submitted program) must match the specified output format, not only the correctness of the implemented approach.
For example, even a missing space or an empty line will be treated as an incorrect answer.
In addition, if the implemented algorithm is improper from the perspective of time or space complexity, the program will be judged incorrect because of exceeding the time or memory limit even if the results are correct.
For example, if a problem that must be solved using a binary search (time complexity is $O(\log n)$) is solved using a linear search (time complexity is $O(n)$), the program will be judged incorrect because of exceeding the time limit.

These specifications, constraints, and evaluation criteria are critical difficulties in solving programming problems.

\section{Research Question}
\label{sec:research-question}

Recent LLMs for source code are considered highly capable of solving introductory programming problems.
This may imply that they have the ability to understand problem descriptions. 
However, instead of interpreting the provided problem description, we suspect that the LLMs are just picking the most pertinent source code from training data based on small fractions of key terms.
Therefore, we assume that even minor changes to the given inputs have a great impact on the code generation performance.

Our primary RQ is, \emph{Do LLMs interpret given programming problems and generate programs with their thought or just retrieve source code from the most relevant problem in training data on the basis of superficial cues?}
We conduct an empirical investigation to address this RQ in the following sections.

\section{Experiments}
\label{sec:experiments}

\begin{figure*}[t]
    \centering
    \includegraphics[width=\linewidth]{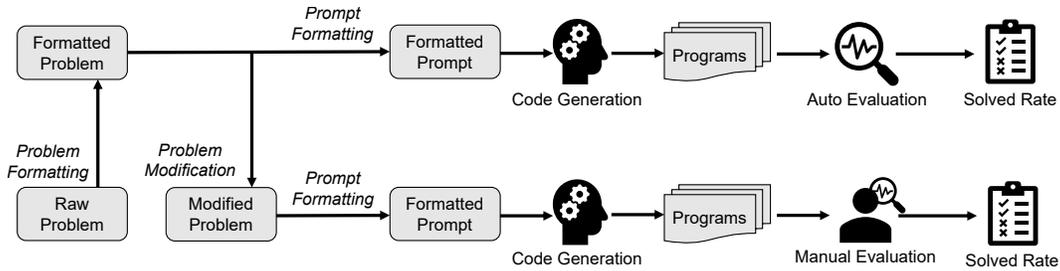}
    \caption{Illustration of the experiments consisting of problem formatting, prompt formatting, code generation, and evaluation.}
    \label{fig:illustration}
\end{figure*}

We carry out two types of experiments to address our RQ.
First, we format the problem descriptions using a set of predefined rules to examine the differences in generated programs.
Second, we modify the problem specifications to observe the effect of the modifications on the generated programs and solved rate.
The experimental framework is illustrated in Figure~\ref{fig:illustration}.

\subsection{Dataset}
\label{sec:experiments:dataset}

We collect a set of programming problems for the experiments, which contains problem descriptions and test cases of the problems.
We use introductory problems provided on AOJ~\citep{watanobe2004aoj}, an online judge system where users solve given programming problems, and the judge system automatically evaluates the submitted programs using test cases~\citep{wasik2018oj}.
AOJ provides approximately 3,000 programming problems ranging from introductory to advanced, along with example test cases and hidden test cases used for evaluation~\citep{watanobe2022aoj}.
Moreover, AOJ provides APIs to access the data and resources\footnote{\url{http://developers.u-aizu.ac.jp/}.}.

We target a course called \textit{Introduction to Programming I} (ITP1)\footnote{\url{https://onlinejudge.u-aizu.ac.jp/courses/lesson/2/ITP1/1}.}, which involves a set of 44 introductory programming problems designed for programming beginners.
AOJ problems are supported in English, Japanese, or both of them. We use the problem descriptions written in English because ITP1 is supported in both English and Japanese, and LLMs are usually more fluent in English.
We exclude four unsuitable problems at the end of the set because they require interpreting an image given in the problem description.
Therefore, there are 40 programming problems in the final target dataset.

\subsection{Models}
\label{sec:experiments:models}

Many Transformer-based~\citep{vaswani2017transformer} autoregressive LLMs trained on source code have been released in the past year, such as Codex~\citep{chen2021codex}, AlphaCode~\citep{li2022alphacode}, PolyCoder~\citep{xu2022polycoder}, CodeGen~\citep{nijkamp2022codegen}, and InCoder~\citep{fried2022incoder}.
Most recently, GPT-3.5 series models\footnote{\url{https://platform.openai.com/docs/models/gpt-3-5}.}, such as InstructGPT~\citep{ouyang2022instructgpt} and ChatGPT, and GPT-4~\citep{openai2023gpt-4}, which are based on Codex, have attracted great attention in not only academia but also industry.

For the experiments, we primarily use Codex and CodeGen models since both are available and are highly capable of code generation in Python.
For comparison, we also use two latest and great capable models from GPT-3.5 series models, InstructGPT and ChatGPT, in the problem formatting experiment.
We consider the other models, such as PolyCoder and InCoder, unsuitable in this work since their sizes (2.7B and 6.7B, respectively) are relatively small compared with Codex and CodeGen (12B and 16B, respectively).
Furthermore, although there have been announcements of other highly capable models, such as AlphaCode and GPT-4, unfortunately, we are unable to access them.

\subsubsection{Codex}
\label{sec:experiments:models:codex}

Codex~12B~\citep{chen2021codex} is a GPT-3-based~\citep{brown2020gpt3} LLM fine-tuned on massive source code publicly available on GitHub and further fine-tuned on 10,000 programming problems.
Since the model is trained on a large number of programming problems, it has quite high capability of solving programming problems, according to the result of the benchmark against APPS~\citep{hendrycks2021apps}, which is designed to evaluate the performance and capability of code generation tasks.

Codex has been used not only for solving programming problems but also in other various tasks, such as solving mathematical problems~\citep{drori2021algebra, tang2021probstat, drori2021math}, fixing bugs~\citep{prenner2021repair, pearce2021securitybugs}, and generating programming exercises~\citep{sarsa2022exercise}.

Codex had offered two models (engines) as API for the limited beta at the time: \texttt{code-davinci-002} and \texttt{code-cushman-002}\footnote{\url{https://platform.openai.com/docs/models/codex}.}.
Although \texttt{code-cushman-002} is much faster, we choose \texttt{code-davinci-002} for the experiments because it is more capable.

\subsubsection{CodeGen}
\label{sec:experiments:models:codegen}

CodeGen~\citep{nijkamp2022codegen} is a Transformer-based~\citep{vaswani2017transformer} autoregressive LLM, incorporating a conversational paradigm, a system that responds to the natural language provided by a user at each turn, called a multi-turn conversation.

CodeGen was firstly pre-trained on the Pile~\citep{gao2021thepile}, a large natural language text corpus, and fine-tuned on two self-collected source code datasets.
\texttt{CodeGen-NL} is the pre-trained model; \texttt{CodeGen-Multi} is the fine-tuned model trained on a multilingual source code dataset consisting of six programming languages such as C, Python, and Java; and \texttt{CodeGen-Mono} is the fine-tuned model further trained on a monolingual source code dataset consisting of only Python.
However, CodeGen has not been fine-tuned on any programming problem datasets, unlike Codex.

The family of CodeGen models varies in the size of parameters with 350M, 2.7B (2B), 6.1B (6B), and 16.1B (16B)%
\footnote{Model names are truncated to the decimal point of the actual size of parameters: 2.7B is 2B, 6.1B is 6B, and 16.1B is 16B.}%
, and the models are publicly available on the Hugging Face Hub%
\footnote{\url{https://huggingface.co/models?search=salesforce+codegen}.}.
Since the experiments target Python~3, we choose \texttt{CodeGen-16B-Mono}%
\footnote{\url{https://huggingface.co/Salesforce/codegen-16B-mono}.}%
, which has the best capability for generating Python programs among the family models.

\subsubsection{GPT-3.5 Models}
\label{sec:experiments:models:gpt-3.5}

GPT-3.5\footnote{\url{https://platform.openai.com/docs/model-index-for-researchers}.} models are a family of LLMs that are based on the GPT-3 architecture, especially Codex.
From the GPT-3.5 models, we use InstructGPT and ChatGPT.

\paragraph{InstructGPT}

InstructGPT~\citep{ouyang2022instructgpt} is trained to follow user instructions and provide detailed responses using an RLHF technique to align with the user's intentions.
This ability has many potential applications, such as in customer service, virtual assistants, and education.
Moreover, its ability to learn from corrective feedback can help reduce errors, thus improving the accuracy of its responses and enhancing the correctness and readability of the generated programs.

OpenAI offers several InstructGPT-based models (engines) through pay-as-you-go APIs, especially the engines \texttt{text-davinci-002} and \texttt{text-davinci-003}.
The APIs of InstructGPT cost \$0.02 for each 1,000 tokens of prompts and completions (i.e., the given problem description and the corresponding generated programs).
In this work, we use the most recent and capable engine, \texttt{text-davinci-003}, which is an improved version of \texttt{text-davinci-002}.

\paragraph{ChatGPT}

ChatGPT\footnote{\url{https://openai.com/blog/chatgpt/}.} is a dialogue-based LLM which interacts with users conversationally.
It is a GPT-3.5 model optimized for conversation and, thus, a sibling model to InstructGPT.~\citep{ouyang2022instructgpt}.
ChatGPT has improved its ability to answer follow-up questions, admit its mistakes, challenge incorrect premises, and reject inappropriate requests.

OpenAI has opened a service to use ChatGPT on GUI, ChatGPT UI\footnote{\url{https://chat.openai.com/chat}.}, which supports GPT-3.5 and GPT-4.
It was initially launched as a free preview, but then OpenAI introduced a paid plan, ChatGPT Plus\footnote{\url{https://openai.com/blog/chatgpt-plus}.}.

The ChatGPT-based engines are offered as pay-as-you-go APIs named \texttt{gpt-3.5-turbo}.
Although the engine is automatically updated with the latest model iteration, OpenAI also offers snapshot engines that will not receive any updates.
The APIs of ChatGPT cost \$0.002 for each 1,000 tokens, which is more than 10 times cheaper than InstructGPT-based engines.
In this work, we use \texttt{gpt-3.5-turbo-0301}\footnote{\url{https://platform.openai.com/docs/models/gpt-3-5}.}, which is a snapshot version of \texttt{gpt-3.5-turbo} from March 1st, 2023, for reproducibility.

\subsubsection{Generation Configuration}
\label{sec:experiments:models:config}

Throughout the experiments, we ask the LLMs to generate programs in zero-shot setting, for which we do not provide any examples to demonstrate the expected inputs and outputs.

We target Python~3 programming language in this work because all the target LLMs are most capable in Python.

We set the $temperature$ parameter, which determines creativity, to 0.8 for all code generation in this work, following the Codex experiments~\citep{chen2021codex}
Although a $temperature$ of zero can produce a mostly deterministic program, a certain amount of creativity is required for evaluating the sensitivity (robustness) to prompts.
Also, the maximum number of tokens to generate is set to 256.

For only ChatGPT, we set the system instruction to \texttt{"Solve the following programming problems in Python. You only need to output solution source code in a code block without any explanations in natural language."} before each code generation.
Since ChatGPT is a dialogue-based model unlike the completion-based models such as Codex and InstructGPT, it often generates excessive detailed explanations about the solution approach in natural language and sometimes show pseudo-code before generating a solution code block, when simply given a problem description.
To suppress the generation of natural language explanation and to generate only solution source code in Python instead, we use the system instruction that ChatGPT officially supports in API to control the model's behavior throughout the conversation.

Furthermore, for only ChatGPT, we append post-processing to remove backticks if they exist in the first and last lines in the generated programs. Since we ask the model to generate in a code block, the output of ChatGPT is generally constructed as a code block in Markdown format, which encloses source code with three backticks.

\subsection{Evaluation}
\label{sec:experiments:evaluation}

\subsubsection{Evaluation Criteria}
\label{sec:experiments:evaluation:criteria}

We use two quantitative evaluation metrics to evaluate code generation for the experiments of solving programming problems: solved rate and average solved rate.

The solved rate refers to the proportion of generated programs that successfully solve the given problem, and it focuses on evaluating how well the model can generate correct programs for each problem. The solved rate is defined as Formula~\ref{formula:solved-rate}, where $P_i$ is the number of correct programs and $F_i$ is the number of incorrect programs.
\begin{equation}
    \text{Solved Rate} = \frac{P_i}{P_i + F_i}
    \label{formula:solved-rate}
\end{equation}

The average solved rate is the average of the solved rate for all programming problems in our dataset. The average solved rate is defined as Formula~\ref{formula:average-solved-rate}, where $n$ is the number of programming problems ($n=40$ in this work).
\begin{equation}
    \text{Average Solved Rate} = \frac{1}{n}\sum_{i}^{n} \frac{P_i}{P_i + F_i}
    \label{formula:average-solved-rate}
\end{equation}

There are several choices for evaluation metrics in code generation tasks:
Exact Match, BLEU~\citep{papineni2002bleu}, CodeBLEU~\citep{ren2020codebleu}, test case average and strict accuracy~\citep{hendrycks2021apps}, success rate~\citep{kulal2019spoc}, pass@$k$~\citep{chen2021codex}, $n@k$~\citep{li2022alphacode}, etc.
In particular, pass@$k$ and $n@k$ are popular metrics in recent works for evaluating datasets that target tasks of solving programming problems.
However, the commonly used metrics focus on evaluating whether the model can solve a problem, rather than evaluating how many generated samples solve each given problem.
Since our goal is to measure the models' robustness to different prompts, these metrics are not the most appropriate for this work.

\subsubsection{Execution Environment}
\label{sec:experiments:evaluation:environment}

LLMs learned from various types of source code on GitHub have the risk of generating malicious or vulnerable programs, and so users are strongly discouraged from automatically executing the generated programs on the host computer without understanding them~\citep{chen2021codex, pearce2021security}.
Therefore, we prepare an isolated sandbox environment that does not harm the host computer to execute the generated programs safely.

\subsection{Formatting Problem Descriptions}
\label{sec:experiments:formatting}

Even with the same problem specifications, making slight modifications to the problem descriptions might significantly impact the generated programs.
Therefore, we define several formatting rules for the problem descriptions; these rules are inspired by various public datasets, such as HumanEval~\citep{chen2021codex}, CodeContests~\citep{li2022alphacode}, and APPS~\citep{hendrycks2021apps}, aiming to benchmark text-to-code generation tasks.
The formatting can be divided into two consecutive phases: \textit{problem formatting} and \textit{prompt formatting}.

\subsubsection{Problem Formatting}
\label{sec:experiments:formatting:problem}

In the problem formatting phase, we format the problem descriptions to explicitly represent the problem specifications.
First, we parse the raw problem description written in HTML into plain texts, and second, we manually modify the plain texts into various defined formats as follows.
The modifications only change the superficial formats, and do not change any problem specifications.
Therefore, we refer to these modifications as \textit{superficial modifications}.

\begin{itemize}
    \item Raw HTML.
    \item Parsed plain.
    \item AlphaCode-inspired.
    \item APPS-inspired.
    \item Fully Markdown-formatted.
\end{itemize}

We list the example problem descriptions of the problem ITP1\_1\_C for each formatting in Appendix~\ref{sec:appendix:formatting}.

\paragraph{Raw HTML}

This format is the original problem description written in HTML, downloaded from AOJ.
Figure~\ref{fig:raw-problem} shows an example of this format.

\paragraph{Parsed Plain}

This format is obtained by automatically removing HTML tags and converting symbols from the Raw HTML format's text to unformatted plain text.
We use the Beautiful Soup\footnote{\url{https://www.crummy.com/software/BeautifulSoup/bs4/doc/}.} library in Python to automatically parse the HTML texts.
Figure~\ref{fig:parsed-problem} shows an example of this format.

\paragraph{AlphaCode-Inspired}

This format is manually modified by us, referring to the prompts used in AlphaCode~\citep{li2022alphacode}.
Figure~\ref{fig:alphacode-problem} shows an example of this format.

In this format, the programming language (i.e., Python~3) is specified first, and then an instruction is provided to generate a solution to the given problem.
Originally, in the AlphaCode's prompts, the ratings and tags for a given problem were specified.
However, we omit them in this work since the programming problems we use declare neither ratings nor tags.
Furthermore, numbers, variable names, and mathematical symbols are surrounded by the \$ symbol and represented in the \TeX\, format, and each line is separated by an empty line.

\paragraph{APPS-Inspired}

Similar to the AlphaCode-inspired format, this format is manually modified by us, referring to the problem descriptions provided by APPS~\citep{hendrycks2021apps}.
Figure~\ref{fig:apps-problem} shows an example of this format.

The major differences from the AlphaCode-inspired format are that (1) a section name is emphasized by surrounding it with five hyphens, and (2) no instruction sentence is provided.
However, same as the AlphaCode-inspired format, numbers, variable names, and mathematical symbols are represented in the \TeX\, format.

\paragraph{Fully Markdown-Formatted}

This format is a hybrid inspired by the AlphaCode, APPS, and Markdown formats.
Figure~\ref{fig:markdown-problem} shows an example of this format.

Following the Markdown format, a section name is prefixed with the \texttt{\#} symbol to indicate the hierarchy of the section.
Main sections are prefixed with one \texttt{\#} symbol, and subsections are prefixed with two \texttt{\#} symbols.
Moreover, the examples of input and output cases are represented as code blocks, placing triple backticks \texttt{\textasciigrave\textasciigrave\textasciigrave} before and after the source code.
As usual, numbers, variable names, and mathematical symbols are represented in the \TeX\, format.

\subsubsection{Prompt Formatting}
\label{sec:experiments:formatting:prompt}

In the prompt formatting phase, we further format the formatted problem description to construct a prompt.
A prompt is an input text for the LLMs. Generally, completion-based LLMs like Codex predict and generate the next tokens for the given prompts, whereas dialogue-based LLMs like ChatGPT are a bit different. 

We define the following formatting rules to construct a prompt, referring to the officially proposed best practices to improve the code generation of Codex\footnote{\url{https://platform.openai.com/docs/guides/code/best-practices}}.
Figure~\ref{fig:prompt} shows an example of a formatted prompt.

\begin{itemize}
    \item Specifying the programming language (i.e., Python~3).
    \item Providing as a docstring or comments.
    \item Appending instruction commands, an instruction sentence, or not.
    \item Defining a \texttt{solve()} function or not.
    \item Providing single import, multiple imports, or not.
    \item Inserting, removing, or keeping empty lines.
\end{itemize}

\paragraph{Specifying Programming Language}

Codex and CodeGen support not only Python but also many other programming languages.
Especially for Codex, the model can generate any programs in a particular programming language by specifying the name of the target programming language.
Although the target model of CodeGen is fine-tuned on the Python dataset and specialized for Python, we suspect that explicitly specifying the target programming language can help the model generate more accurate programs since the model is pre-trained on many programming languages.
Therefore, throughout the experiments in this work, in the first line of the prompt, we append a comment specifying the Python~3 programming language (i.e., \texttt{\# Python 3}).

\paragraph{Providing as a Docstring or Comments}

The problem description should be formed as a docstring or comments since the prompt needs to be a part of the Python program.
One recommended best practice, which is announced by Codex's official document, is to use docstrings instead of comments because this can produce higher-quality results.
Therefore, we investigate why docstrings are better in Codex and which option is suited to CodeGen.

\paragraph{Appending Instruction Commands or a Sentence}

The prompts used in AlphaCode have an instruction sentence to solve the given problem, so just giving the problem description may cause the LLMs to struggle to decide what to do next.
Therefore, we append two types of instructions to ask the LLMs to start solving the given problem: (1) instruction commands and (2) an instruction sentence.

For instruction commands, to indicate the range of the problem description, we append \texttt{\textless START PROBLEM DESCRIPTION\textgreater} at the beginning and \texttt{\textless END PROBLEM DESCRIPTION\textgreater} at the end of the prompt.
For an instruction sentence, to indicate the end of the problem description, we append \texttt{"Write code to solve the above problem."} at the end of the prompt.

\paragraph{Defining a \texttt{solve()} Function}

Another best practice mentioned in Codex is to place a description of the function inside the function as a docstring.
This can help Codex understand what we want the function to accomplish.
Therefore, we provide the problem description as a docstring by defining the function \texttt{solve()}.

However, different from the case illustrated in the best practice, this format may not be effective in our experiments for the following reasons:
(1) We ask the LLMs to generate an entire program to solve the given problem, rather than a function body with a specific purpose.
(2) We can provide the LLMs with very little additional information since the defined function name (i.e., solve) is too common.

\paragraph{Providing a Single or Multiple Imports}

According to the best practices, providing the necessary libraries can be beneficial.
In this experiment, we examine two types of imports, namely, (1) a single import and (2) multiple imports.

For a single import, only the \texttt{math} library is provided in the prompt, which is surely required in some problems.
For multiple imports, several commonly used libraries are provided in the prompt, including \texttt{os}, \texttt{re}, \texttt{sys}, and \texttt{math}.
Although the libraries can be used in some solution approaches, they are not essential libraries except the \texttt{math} library.
Because problems vary in their solution approaches, even if unnecessary libraries are imported in the prompts, we expect the LLMs to ignore libraries when they are not necessary for the solution of a given problem.

\paragraph{Inserting or Removing Empty Lines}

We also investigate the sensitivity of the LLMs to the empty lines of the problem descriptions because suitable breaks between sentences enhance text readability.
Accordingly, we conduct two format types: (1) inserting an empty line between each sentence and (2) removing all empty lines from the problem description.

\subsection{Modifying Problem Descriptions}
\label{sec:experiments:modification}

There is a concern that the LLMs memorize typical programming problems and generate the corresponding most relevant source code, which is contained in the training data.
We suspect that the performance for the same problem can be worsen if the LLMs simply memorize problems.

Therefore, we conduct four experiments where we alter the problem description to assess whether the LLMs understand the details of the given problem description.
We modify not only the superficial formatting but also the information related to problem specification.

In the following experiments, generating correct programs is not mandatory because the experiments aim to determine whether the LLMs can generate different programs from the original one to follow the modification.
Furthermore, even if the modified problem specification is uncommon and difficult, we expect the LLMs to generate programs that attempt to solve the modified problem.

\subsubsection{Randomized Variable Names}
\label{sec:experiments:modification:variables}

According to the qualitative evaluation of our early experiments, we observed that the generated programs often use the same variable names defined in the problem description.
This suggests that the LLMs may be able to catch the variable names from the problem description.
However, almost only typical variable names are used in the problem description, such as $x$, $y$, $W$, and $H$.
Therefore, we suspect that the variable names used in the problem description might be the key to retrieving the most relevant solution programs from the training data.

This experiment aims to investigate
(1) how an LLM catches the variable names defined in the problem description and
(2) whether the generated program uses the defined variable names or defines new ones.
We conduct three types of randomization: \textit{UUID variables}, \textit{shuffled variables},  and \textit{ABC variables}.

\paragraph{UUID Variables}

To ensure that the defined variable names do not appear in the training data and the LLM has never seen them, we anonymize all variable names in the problem descriptions with universally unique identifiers (UUID)~\citep{leach2005uuid}.
For practical purposes, UUIDs are unique, with an extremely low probability of duplication.
Moreover, to ensure that the LLM can use the anonymized variable names as actual ones in the Python program, we only use the UUIDs that start with an alphabet, and hyphens are removed.

\begin{figure*}[ht]
    \centering
    \includegraphics[width=0.8\linewidth]{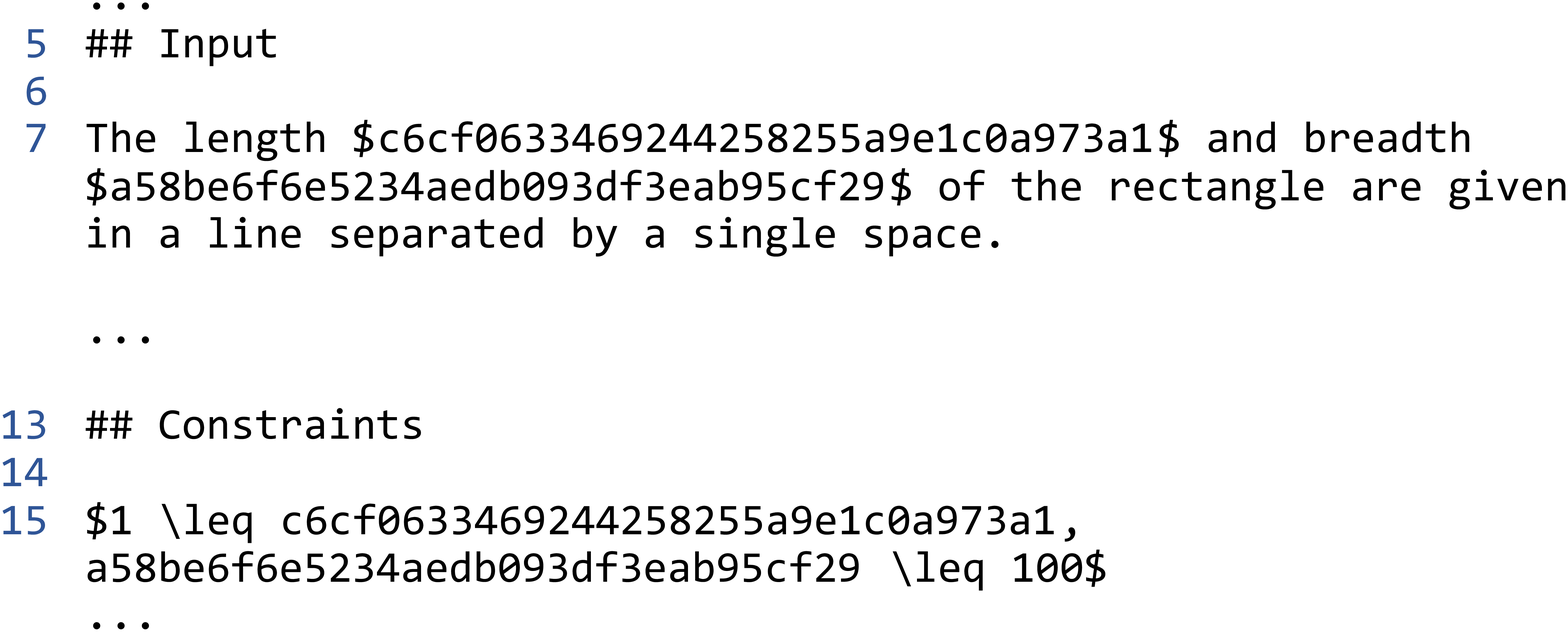}
    \caption{A partial example of a fully Markdown-formatted problem (ITP1\_1\_C), with anonymized variables $a$ and $b$.}
    \label{fig:variables-anonymized-problem}
\end{figure*}

Figure~\ref{fig:variables-anonymized-problem} shows a partial example of a problem description, anonymizing variables $a$ and $b$ with $6cf0633469244258255a9e1c0a973a$ and $a58be6f6e5234aedb093df3eab95cf$, respectively.

\paragraph{Shuffled Variables}

Since the UUIDs are unique and uncommon, the experiment setting might be much more difficult than we assumed.
Furthermore, since the UUIDs are long, 32 characters excluding hyphens, an anonymized variable name can be tokenized into too many tokens and thus can be challenging to treat as one variable name.
Using the official tokenizer of Codex\footnote{\url{https://beta.openai.com/tokenizer}.}, we confirmed that UUID variables are tokenized into 20 tokens on average.

To simplify the experimental setting, in this randomization, the original variables defined in the problem description are shuffled.
For example, if $H$ originally indicates the height and $W$ the width, they will indicate the width and height, respectively.

This randomization especially aims to determine whether the common variable names affect the LLM in situations where the shuffled variable names are guaranteed to be common, being originally used, but the usage is different and unusual.

\paragraph{ABC Variables}

In this randomization, the original variables are replaced with variables named $a, b, c, \cdots$.
If the original variables are already $a, b, c$, then the variables will be replaced with $x, y, z$.

We expect that even if the solved rate decreases, the decreasing rate is better than the shuffled variables.
This randomization removes the superficial information; but does not intend to confuse the LLM about the meaning of variable names.

\subsubsection{Anonymized Output Strings}
\label{sec:experiments:modification:outputs}

Similar to Section~\ref{sec:experiments:modification:variables}, we doubt whether the LLM generates the specified output strings by interpreting the problem description.
We suspect that the LLM uses the same output strings present in the training data because the specified output strings in the problem descriptions are also common.
Therefore, this experiment aims to investigate whether the LLM catches the output strings specified in the problem description.

We anonymize all output strings specified in the problem description using UUIDs.
Unlike the specification of UUIDs used in Section~\ref{sec:experiments:modification:variables}, for this experiment, the anonymized output strings may start with a number, as they are assumed to be defined as string literals rather than variable names.

\subsubsection{Rewording Synonyms}
\label{sec:experiments:modification:synonyms}

We suspect that the LLM depends on some specific words appearing in the problem description.
For example, we suspect that the solved rate can be decreased by simply rewording synonyms, owing to the bias of the words that often appears in the training data, or because the LLM cannot find the meanings of certain words that humans would consider almost the same with no difference in meaning.

This experiment investigates whether synonyms change the solved rate.
We replace some words with synonyms to an extent that does not adversely affect the meaning of the problem specification.
For example, \texttt{"ascending order"} is replaced with \texttt{"increasing order"}, and \texttt{"print"} is replaced with \texttt{"output"}.

\subsubsection{Inverse Problem Specification}
\label{sec:experiments:modification:inverse}

The LLMs can usually solve a problem with a high solved rate if the problem specification is typical and the problem description is clearly stated.
We consider that the LLM memorizes the solution approach or the algorithm from some phrases in the problem description.
Therefore, we conduct an experiment of semantic modification to determine
(1) whether the LLM can follow the modification of problem specification even if it is uncommon or difficult and
(2) how sensitive the LLM is to the modification.

We perform common and uncommon modifications to make the problem specification opposite but still valid.
For example, modifying the sentence from \texttt{"output values in ascending order"} to \texttt{"output values in descending order"} is a common modification because both settings are typical for introductory programming with almost the same difficulty.
On the other hand, modifying the sentence from \texttt{"determining whether a circle is arranged inside a rectangle"} to \texttt{"determining whether a rectangle is arranged inside a circle"} is an uncommon modification because the new setting is a bit more difficult than the original, but it is still valid and can be solved.

\section{Results and Discussion}
\label{sec:results}

\subsection{Formatting Problem Descriptions}
\label{sec:results:formatting}

In the experiments in this section, each LLM generates 100 programs for each of the 40 problems, and the judge system automatically validates the results.

\subsubsection{Problem Formatting}
\label{sec:results:formatting:problem}

\begin{table}[ht]
\begin{center}
\begin{minipage}{0.8\linewidth}
    \caption{Average solved rate (\%) for each type of problem formatting, generating 100 programs for each of the 40 problems. The best formatting in each model is shown in \bn{bold}.}
    \label{table:problems-result}
    \begin{tabular*}{\linewidth}{lrrrr}
        \toprule
        Formatting & CodeGen & Codex & InstructGPT & ChatGPT \\
        \midrule
        Raw HTML & 10.9 & 30.9 & \bn{75.9} & \bn{90.1} \\
        Parsed plain & 9.2 & 34.7 & 73.5 & 89.0 \\
        AlphaCode-inspired & 9.7 & 35.2 & 72.0 & 88.7 \\
        APPS-inspired & \bn{11.7} & 39.3 & 73.7 & 88.6 \\
        Fully Markdown-formatted & 9.9 & \bn{39.9} & 74.5 & 89.0 \\
        \midrule
        Average & 10.3 & 36.0 & 73.9 & 89.1 \\
        Variance & 1.01 & 13.61 & 2.04 & 0.36 \\
        \bottomrule
    \end{tabular*}
\end{minipage}
\end{center}
\end{table}

Table~\ref{table:problems-result} shows the average solved rate for each type of problem formatting, conducted on four models, CodeGen, Codex, InstructGPT, and ChatGPT.

On average, Codex shows three times better performance than the CodeGen.
Codex's successor, InstructGPT, doubles the average solved rate, and ChatGPT shows an even greater improvement.
This large performance difference reflects the effect of fine-tuning for programming problems, as Codex is fine-tuned whereas CodeGen is not.

For Codex, the average solved rate is improved by simply parsing the Raw HTML problems into plain text, and the fully Markdown-formatted problems show the highest average solved rate with a $9.0\%$ performance improvement compared with the Raw HTML format.
In Codex, the more explicitly formatted problem descriptions enhance the average solved rate of code generation.
Accordingly, the effective formatting rules in Codex are (1) inserting an empty line between each sentence, (2) separating section names, and (3) representing example input and output cases as code blocks.
However, we did not find a clear improvement in the other models.

Conversely, the improved performance in problem formatting in Codex suggests that the model is highly influenced by the superficial modification of the problem description and does not understand the problem description in detail, since problem formatting affects only the problem description while keeping the problem specification the same.

For InstructGPT and ChatGPT, the Raw HTML format yields the best performance, but it is not the best format in both CodeGen and Codex.
Furthermore, the variance in the average solved rate is decreasing as the models evolve, except in CodeGen since its performance is extremely low.
Although there still remains a high possibility of data leakage, robustness to superficial modifications may increase, as observed in Codex.

\begin{table}[ht]
\begin{center}
\begin{minipage}{0.8\linewidth}
    \caption{Average number of lines, comment lines, empty lines, and actual processing lines in programs generated for fully Markdown-formatted problems. Values in parentheses indicate the proportion. $\text{Lines} = \text{CommentLines} + \text{EmptyLines} + \text{ProcessLines}$.}
    \label{table:lines-result}
    \begin{tabular*}{\linewidth}{lrrrr}
        \toprule
        Model & Lines & CommentLines & EmptyLines & ProcessLines \\
        \midrule
        CodeGen & 30.8 & 11.7 (38.0\%) & 8.2 (26.6\%) & 10.9 (35.4\%) \\
        Codex & 19.1 & 2.1 (11.0\%) & 6.0 (31.4\%) & 11.0 (57.6\%) \\
        InstructGPT & 12.4 & 1.1 (8.9\%) & 2.8 (22.6\%) & 8.5 (68.6\%) \\
        ChatGPT & 10.2 & 0.1 (1.0\%) & 1.9 (18.6\%) & 8.1 (79.4\%) \\
        \midrule
        Users' Solutions & 15.7 & 0.4 (2.5\%) & 2.9 (18.5\%) & 12.4 (79.0\%) \\
        \bottomrule
    \end{tabular*}
\end{minipage}
\end{center}
\end{table}

For the characteristics of the generated programs, Table~\ref{table:lines-result} shows the average number of lines of generated programs, along with the average of users' solution programs for reference.
\textit{Lines} indicates the pure number of lines, also known as lines of code (LOC), which is simply calculated as the number of lines in the generated program.
\textit{CommentLines} indicates the number of comment lines, including lines that form docstrings (i.e., multi-lined comments).
\textit{EmptyLines} indicates the number of empty lines, including lines that only contain white spaces.
\textit{ProcessLines} indicates the number of actual processing lines, that is, lines that are neither comment or empty lines.
However, empty lines of comments or docstrings are counted as comment lines, not empty lines.

As the model evolves from Codex to ChatGPT, the number of lines becomes smaller for all metrics, indicating that the generated program is becoming more concise.
Although the ProcessLines is equivalent in CodeGen and Codex, CodeGen generates much longer comment lines and empty lines than Codex.
CommentLines and EmptyLines can be decreased by simply removing the comment and empty lines, whereas the decrease in ProcessLines suggests that the program itself becomes more concise.

In ChatGPT, almost no comments are generated.
As the number of comment lines in users' solutions is also very small, we consider that ChatGPT might be designed not to generate excessive comments in the programs.
However, this might be influenced by the system instruction that we instructed ChatGPT to generate only a code block and to suppress generating natural language explanations, as introduced in Section~\ref{sec:experiments:models:config}.
Although we intended to suppress generating explanations outside of a code block, not comments in the programs, this may cause the suppression of comments in natural language.

Since moderate comments and empty lines are essential for readability, we cannot directly conclude that these decreases in the number of lines in the generated programs indicate the capability of the LLMs.
However, an interesting observation is that the SOTA LLMs attempt to make the programs concise.

\subsubsection{Prompt Formatting}
\label{sec:results:formatting:prompt}

\begin{table}[ht]
\begin{center}
\begin{minipage}{0.95\linewidth}
    \caption{Results for each type of prompt formatting, generating 100 programs for each of the 40 problems. ASR (\%) indicates the average solved rate, ACP (\%) indicates the average comment proportion in the generated programs, and AL indicates the average number of lines of generated programs including empty lines. The best ASR in each model is shown in \bn{bold}.}
    \label{table:prompts-result}
    \begin{tabular*}{\linewidth}{lrrrrrr}
        \toprule
        & \multicolumn{3}{c}{Codex} & \multicolumn{3}{c}{CodeGen} \\
        \cmidrule{2-4}\cmidrule{5-7}
        Formatting & ASR & ACP & AL & ASR & ACP & AL \\
        \midrule
        Docstring & \bn{39.9} & 10.9 & 19.1 & 9.9 & 37.9 & 30.8 \\
        Docstring + appending imports & \bn{39.9} & \NA & \NA & 9.9 & \NA & \NA \\
        Comments & 27.9 & 41.4 & 53.0 & 6.4 & 79.4 & 49.3 \\
        Comments + appending imports & 27.9 & \NA & \NA & 6.4 & \NA & \NA \\
        \midrule
        Instruction command (Docstring) & 19.4 & 23.9 & 36.7 & 6.8 & 40.6 & 32.3 \\
        Instruction command (Comments) & 14.0 & 40.4 & 103.5 & 10.1 & 47.2 & 31.3 \\
        Instruction sentence (Docstring) & 28.4 & 14.3 & 22.1 & 7.2 & 42.3 & 32.5 \\
        Instruction sentence (Comments) & 21.4 & 36.3 & 79.3 & 8.3 & 59.6 & 37.5 \\
        \midrule
        Defining \texttt{solve()} & 14.8 & 21.8 & 49.0 & 7.7 & 44.4 & 27.3 \\
        Defining \texttt{solve()} + appending invocation & 26.1 & \NA & \NA & \bn{17.4} & \NA & \NA \\
        \midrule
        Providing single import & 35.9 & 9.6 & 24.3 & 3.0 & 59.6 & 36.0 \\
        Providing multiple imports & 31.0 & 8.3 & 21.5 & 2.7 & 49.6 & 33.4 \\
        \midrule
        Inserting empty lines & 33.7 & 11.2 & 25.4 & 8.6 & 36.6 & 34.6 \\
        Removing empty lines & 30.8 & 18.3 & 27.6 & 7.4 & 42.4 & 30.4 \\
        \bottomrule
    \end{tabular*}
\end{minipage}
\end{center}
\end{table}

Table~\ref{table:prompts-result} shows the average solved rate for each type of prompt formatting, conducted on Codex and CodeGen.
Providing the problem description as a docstring (Docstring) is the baseline because the problem formatting experiments are conducted using this format.

\paragraph{Providing as a Docstring or Comments}

Although providing the problem descriptions as comments do not modify the problem descriptions at all, it significantly decreases the average solved rate in both models.

Our qualitative evaluation revealed that, in this format, most of the incorrect programs do not try to solve the problem, and they continue to write comments instead.
This qualitative observation agrees with our quantitative finding that 41.4\% of lines in the generated programs, excluding the given prompts, are comments when providing in the comment format, whereas only 10.9\% are comments when providing in the docstring format in Codex.
The same issue of increasing comment proportion from the comment format to the docstring format is observed in CodeGen ($37.9\% \rightarrow 79.4\%$).

This results suggest that the worsened performance in the comment format is due to the unclear scope of the problem description.
On the other hand, docstring format can explicitly indicate the beginning and end of the problem description by """ symbols to define the docstring.

\paragraph{Appending Instruction Commands or a Sentence}

The effect of appending instruction commands or an instruction sentence is different in Codex and CodeGen: a significant decrease in Codex and not significant in CodeGen.

As for the comment format in both models, the average comment proportion is reduced by appending instruction commands ($41.4\% \rightarrow 40.4\%$ in Codex and $79.4\% \rightarrow 47.2\%$ in CodeGen) or sentences ($41.4\% \rightarrow 36.3\%$ in Codex and $79.4\% \rightarrow 59.6\%$ in CodeGen).
Accordingly, they help indicate the end of the problem description.

In contrast, in the docstring format, the average comment proportion is increased in both models with instruction command ($10.9\% \rightarrow 23.9\%$ in Codex and $37.9\% \rightarrow 40.6\%$ in CodeGen) and with instruction sentence ($10.9\% \rightarrow 14.3\%$ in Codex and $37.9\% \rightarrow 42.3\%$ in CodeGen).
Accordingly, the docstring format is sufficient to indicate the end of the problem description.
Moreover, our qualitative evaluation showed that many generated programs are affected by the instruction and generate similar comments, such as \texttt{\textless START YOUR CODE\textgreater} and \texttt{\textless END YOUR CODE\textgreater}.

\paragraph{Defining a \texttt{solve()} Function}

Defining a \texttt{solve()} function shows quite the opposite effects in CodeGen and Codex.
With appending invocation as post-processing, CodeGen shows a significant improvement and performs best, whereas Codex shows a significant decrease in performance.

In Codex, defining a \texttt{solve()} function significantly decreases the average solved rate.
Too much emphasis is placed on implementing the defined function, and many generated programs do not invoke the function.
When we append post-processing to invoke the defined function if the generated program does not invoke it, the average solved rate increases by 11.3\% ($14.8\% \rightarrow 26.1\%$).
However, the one with appending invocation is still less accurate than the docstring format.
Although this format can be better to implement a single function definition, it is considered inappropriate if flexibility is required, such as solving programming problems.
This format enforces the model to solve using only one function.

In CodeGen, as opposed to Codex, the average solved rate is significantly improved and shows the best performance when appending function invocation.
This result strongly reflects that CodeGen excels at implementing function definitions rather than a whole program.
However, similar to those in Codex, many generated programs in CodeGen only implement the function body and do not invoke the defined function.
Furthermore, since CodeGen is based on conversational program synthesis, it is considered unsuitable for automatic problem-solving where only the problem description is given.

\paragraph{Providing a Single or Multiple Imports}

We observed that the more the imports provided, the worse the models' performance.
Therefore, we suggest not to provide imports and to leave the entire code generation to the model.

Both CodeGen and Codex are affected by the provided imports, and to use the libraries, they are forced to generate more LOC that are unnecessary for solving the given problem.
Although some problems require the provided imports and others do not, the average solved rate is decreased in both.

We also found that appending imports in front of the generated program as post-processing (\textit{Docstring + appending imports} and \textit{Comments + appending imports}) does not improve the average solved rate.
This shows that no generated programs miss necessary imports, and the models can import the libraries when they are required for the solution.

We conclude that providing imports does not help solve the given problem more accurately even if the problem requires them.

\paragraph{Inserting or Removing Empty Lines}

For inserting and removing empty lines, both formats in both models slightly decrease the solved rate, which demonstrates that LLMs are sensitive even to the empty lines.
Moreover, the results do not support our expectation that inserting empty lines can make the problem description more readable.
However, inserting is better than removing empty lines.
Accordingly, we recommend using an appropriate number of empty lines (i.e., not inserting too much or removing too much) for better code generation.

\subsection{Modifying Problem Descriptions}
\label{sec:results:modification}

In this section, we focus on using Codex in our experiments, as the results of problem and prompt formatting have shown that Codex performs better than CodeGen in solving programming problems.
We manually evaluate the 20 programs generated by Codex for each problem.

\subsubsection{Randomized Variable Names}
\label{sec:results:modification:random-variables}

\begin{table}[ht]
\begin{center}
\begin{minipage}{0.83\linewidth}
    \caption{Results of randomized variable names, with 20 programs generated for each problem.}
    \label{table:randomized-variables-result}
    \begin{tabular*}{\linewidth}
    {lrrrrr}
        \toprule
        & & \multicolumn{2}{c}{Average Solved Rate (\%)} \\
        \cmidrule{3-4}
        Randomization & n & Original & Randomized & Change Rate (\%) & P-value \\
        \midrule
        UUID & 10 & 54.9 & 34.0 & -38.1 & $< 0.05$ \\ 
        Shuffled & 9 & 52.0 & 32.8 & -37.0 & $< 0.05$ \\ 
        ABC & 10 & 54.6 & 42.0 & -23.1 & $< 0.05$ \\ 
        \bottomrule
    \end{tabular*}
\end{minipage}
\end{center}
\end{table}

Table~\ref{table:randomized-variables-result} illustrates that there is a statistically significant difference between the original and randomized variables for all types of randomization, using the Wilcoxon signed-rank test where the significance level is set to 0.05.
This finding suggests that the model heavily relies on the information of the variable names defined in the problem description, although the problem specification is not changed.

Both UUID and shuffled variables show a significant drop in the average solved rate, and the drop caused by UUID variables is larger.
This result indicates that the UUID variables complicate the problem, as we considered in Section~\ref{sec:experiments:modification:variables}. 
However, these decreases in both variables suggest that the larger drop in the UUID variables is not only due to its unique format and the large number of tokens to be tokenized.

Moreover, it is seen that ABC variables also decrease the average solved rate significantly.
Unlike the UUID and shuffled variables, ABC variables are assumed not to distract the model by variable naming and to anonymize the meaning of the variables.
The decrease in ABC variables further suggests that the model relies on the information of the variable names.

However, the use of confusing variable names should be avoided, and the variable names that play some typical roles, such as $H$ for height, $W$ for width, and $r$ for radius, should be used only as such for better code generation.
For future work, we suggest that having many variations of variable names used in the training data may mitigate this issue.

\begin{table}[ht]
\begin{center}
\begin{minipage}{0.68\linewidth}
    \caption{Proportion of generated programs using UUID variables with the same variable names defined in the problem description for each problem, with 20 programs generated for each problem.}
    \label{table:anonymized-variables-result}
    \begin{tabular*}{\linewidth}
    {llrr}
        \toprule
        & & \multicolumn{2}{c}{Using Defined Variables (\%)} \\
        \cmidrule{3-4}
        Problem & Original Variables & Original & Anonymized \\
        \midrule
        ITP1\_1\_C & $a, b$ & 80 & 30 \\
        ITP1\_2\_B & $a, b, c$ & 95 & 20 \\
        ITP1\_2\_D & $H, W, x, y, r$ & 100 & 5 \\
        ITP1\_3\_B & $x, i$ & 60 & 5 \\
        ITP1\_3\_C & $x, y$ & 80 & 0 \\
        ITP1\_3\_D & $a, b, c$ & 85 & 15 \\
        ITP1\_4\_B & $r$ & 75 & 0 \\
        ITP1\_5\_A & $H, W$ & 60 & 5 \\
        ITP1\_5\_B & $H, W$ & 30 & 15 \\
        ITP1\_5\_C & $H, W$ & 60 & 5 \\
        \midrule
        Average & & 73 & 10 \\
        \bottomrule
    \end{tabular*}
\end{minipage}
\end{center}
\end{table}

Additionally, Table~\ref{table:anonymized-variables-result} shows that the programs generated for the modified problems rarely use the anonymized variable names, although many generated programs use the defined variable names in the original problems, and the UUID variables are ensured to be used as actual variable names. 
This implies that the knowledge of the LLM about variable naming refuses to use the anonymized variable names since they are uncommon.
A similar observation is reported in the past study~\citep{troshin2022probing},

\subsubsection{Anonymized Output Strings}
\label{sec:results:modification:outputs}

\begin{table}[ht]
\begin{center}
\begin{minipage}{0.95\linewidth}
    \caption{Results of anonymized output strings, with 20 programs generated for each problem. \textit{Using Unrelated Strings} indicates the proportion of generated programs that attempt to output unrelated strings, which are not specified in the problem description.}
    \label{table:outputs-result}
    \begin{tabular*}{\linewidth}
    {lp{0.2\linewidth}rrr}
        \toprule
        & & \multicolumn{2}{c}{Solved Rate (\%)} \\
        \cmidrule{3-4}
        Problem & Original Strings & Original & Anonymized & Using Unrelated Strings (\%) \\
        \midrule
        ITP1\_1\_A & \textit{"Hello World"} & 93 & 70 & 25 \\
        ITP1\_2\_A & \textit{"$a < b$", "$a > b$", "$a == b$"} & 88 & 45 & 5 \\
        ITP1\_2\_D & \textit{"Yes"}, \textit{"No"} & 22 & 10 & 0 \\
        ITP1\_3\_A & \textit{"Hello World"} & 73 & 30 & 0 \\
        \midrule
        Average & & 69.0 & 38.8 & 7.5 \\
        \bottomrule
    \end{tabular*}
\end{minipage}
\end{center}
\end{table}

Table~\ref{table:outputs-result} shows the target output strings to be anonymized and the change in their solved rates.
Compared with that of the original output strings, the average solved rate is significantly decreased when anonymized.

Our qualitative evaluation indicates that the reason for the decreasing solved rate is the unique and uncommon format of the anonymized strings.
We observed a few cases where the incorrectly generated programs regarded the anonymized output strings as encoded strings and attempted to output by encoding some strings.
For instance, we observed a case of encoding an integer from 0 to 99 using a hash function, SHA-1, using the \texttt{hashlib} library and outputting its hexadecimal letters.
However, we also found that they rarely attempt to output completely unrelated strings that do not appear in the problem description.

These results indicate that the LLM can understand which string is the output string although it often fails to find meaning in the anonymized output strings.

\subsubsection{Rewording Synonyms}
\label{sec:results:modification:synonyms}

\begin{table}[ht]
\begin{center}
\begin{minipage}{0.73\linewidth}
    \caption{Results of rewording synonyms, generating 100 programs for each of $n$ problems. No statistically significant differences were found between any synonyms using the Wilcoxon signed-rank test.}
    \label{table:rewording-synonyms}
    \begin{tabular*}{\linewidth}
    {llrrrr}
        \toprule
        \multicolumn{2}{c}{Rewording} & \multicolumn{2}{c}{Average Solved Rate (\%)} \\
        \cmidrule{1-2}\cmidrule{3-4}
        Original & Replaced & Original & Replaced & $n$ & P-value \\
        \midrule
        \textit{print} & \textit{output} & 74.1 & 74.6 & 12 & 0.4750 \\
        \textit{calculate} & \textit{compute} & 56.4 & 57.6 & 8 & 0.7422 \\
        \textit{integer} &\textit{number} & 69.7 & 70.6 & 7 & 0.9156 \\
        \textit{read} & \textit{receive} & 67.8 & 66.3 & 6 & 0.6875 \\
        \textit{ascending} & \textit{increasing} & 70.5 & 73.0 & 2 & 0.5000 \\
        \textit{convert} & \textit{parse} & 59.0 & 58.5 & 2 & 1.0000 \\
        \bottomrule
    \end{tabular*}
\end{minipage}
\end{center}
\end{table}

Table~\ref{table:rewording-synonyms} shows the result of rewording synonyms.
This result does not meet our expectation that the code generation depends on a particular word in the problem description and rewording it would change the solved rate.
Changing the format would change the solved rate significantly, but not in rewording.

\subsubsection{Inverse Problem Specification}
\label{sec:results:modification:inversion}

\begin{table}[ht]
\begin{center}
\begin{minipage}{\linewidth}
    \caption{Results of inverse problem specification, with 20 programs generated for each problem. \textit{Follows} is the proportion of generated programs that follow the modification using a correct approach. The modified parts are indicated in \textbf{bold}.}
    \label{table:inversion-result}
    \begin{tabular*}{\linewidth}
    {lp{0.21\linewidth} p{0.21\linewidth}rrr}
        \toprule
        & \multicolumn{2}{c}{Specification} & \multicolumn{2}{c}{Solved Rate (\%)} \\
        \cmidrule{2-3}\cmidrule{4-5}
        Problem & Original & Modified & Original & Modified & Follows (\%) \\ 
        \midrule
        ITP1\_2\_B & Output \textit{Yes} if the values are in \textbf{ascending} order, \textit{No} otherwise. & Output \textit{Yes} if the values are in \textbf{descending} order, \textit{No} otherwise. & 71 & 40 & 80 \\
        ITP1\_2\_C & Output the values in \textbf{ascending} order. & Output the values in \textbf{descending} order. & 54 & 35 & 95 \\
        ITP1\_2\_D & Output \textit{Yes} if \textbf{the circle is placed inside the rectangle}, \textit{No} otherwise. & Output \textit{Yes} if \textbf{the rectangle is placed inside the circle}, \textit{No} otherwise. & 22 & 0 & 40 \\
        \midrule
        Average & & & 49 & 25 & 72 \\
        \bottomrule
    \end{tabular*}
\end{minipage}
\end{center}
\end{table}

The results of the inverse problem specification and modifications details (the original and modified sentences) are shown in Table~\ref{table:inversion-result}.
Overall, the average solved rate is decreased in modified problems, although many generated programs could follow the solution approach for the modified version.

For ITP1\_2\_C, which is a problem requiring output values in descending order, the solved rate for the modified problem is significantly decreased from 71\% to 40\%.
The only additional processing required for the modified problem is to set the reverse order argument for the sorting function or to reverse the sorted list.
Since the modified problem is also quite common, its difficulty is considered almost equivalent to the original version.
Assuming the LLM comprehended the original problem, the modified problem is expected to have a similar solved rate.

In fact, we ascertain that the LLM can adapt to the modifications and strive to address the modified problem accordingly.
Most of the generated programs employ correct solution approaches, such as attempting to reverse the given list of values, but some non-critical errors cause the wrong answer.
The primary errors in the generated programs include receiving unnecessary inputs, outputting unnecessary strings, executing similar processes multiple times, and using incorrect output format (e.g., missing newlines and spaces).

A similar issue arises in the modified problem requiring the determination of whether the values are in descending order (Inversion of ITP1\_2\_B).
In this case, we find 3 out of 20 programs where the \textbf{generated programs ignore the modification and solve the original problem}.
In other words, the LLM can generate programs implementing exactly the opposite solution approach for the given problem.

As for the uncommon modification (Inversion of ITP1\_2\_D), which is the problem requiring the determination of whether a rectangle is arranged inside a circle, none of the generated programs successfully solve the modified problem.
For the modified problem, the expected solution is to calculate the distance from the circle's center to each of the rectangle's four corners and verify that the distances are less than or equal to the circle's radius.
It is noteworthy that 8 out of 20 generated programs attempt to adapt to the modification and calculate the distances required to solve.
However, although it is expected that the increased difficulty of the modified problem can lead to a lower solved rate, we expected that the modified problem should also be solved.
Furthermore, like the case with the inversion of ITP1\_2\_D, two generated programs ignored the modification and solved the original problem instead.

These results suggest that code generation is affected by statistical biases, such as typical and common programming problems.
We also conclude that, despite their ability to solve problems, LLMs can generate programs that target the exact opposite specification or output unnecessary source code.

\section{Related Work}
\label{sec:related-work}

Various code-related tasks, such as code generation and code understanding, are handled by LMs.
Examining their capabilities and limitations have attracted research interest in recent years.

\subsection{LMs for Code}
\label{sec:related-work:lms}

Here, we introduce the overview of LMs trained on source code aiming to perform code-related tasks.
There are three main types of Transformer-based~\citep{vaswani2017transformer} LMs: encoder models, decoder models, and encoder-decoder models.

\paragraph{Encoder Models}

Encoder models such as BERT~\citep{devlin2019bert} and RoBERTa~\citep{liu2019roberta} are models that use only the encoder part of the Transformer.
Several BERT-based models for code, such as CodeBERT~\citep{feng2020codebert}, CuBERT~\citep{kanade2020cubert}, and SynCoBERT~\citep{wang2021syncobert}, were proposed.
They excel at downstream tasks requiring an understandings of source code, such as clone detection, defect detection, code summarization, and code search.
TreeBERT~\citep{jiang2021treebert} uses not only the source code and natural language but also the syntactic structure of source code, based on abstract syntax tree (AST).
GraphCodeBERT~\citep{guo2021graphcodebert} also uses graph data flow between variables, further transformed from the AST.

\paragraph{Decoder Models}

Decoder models such as GPT-3~\citep{brown2020gpt3} and PaLM~\citep{chowdhery2022palm} are auto-regressive models that use only the decoder part of the Transformer.
Many auto-regressive LLMs for code, such as Codex~\citep{chen2021codex}, PolyCoder~\citep{xu2022polycoder}, CodeGen~\citep{nijkamp2022codegen}, InCoder~\citep{fried2022incoder}, and PanGu-Coder~\citep{christopoulou2022pangucoder}, were proposed in the past year.
While the number of parameters of ChatGPT and GPT-4~\citep{openai2023gpt-4}, which are also trained on source code but private, is considered huge, PaLM-Coder~\citep{chowdhery2022palm}, which has 540B parameters, is currently the largest LM trained on source code.

\paragraph{Encoder-Decoder Models}

Encoder-decoder models, also known as sequence-to-sequence (seq2seq) models, such as BART~\citep{lewis2020bart} and T5~\citep{raffel2020t5}, use both encoder and decoder parts of the Transformer.
These models excel in text-to-text conversion tasks such as text-to-code generation, code-to-text summarization, code-to-code translation, and discriminative tasks, such as program repair, clone detection, and defect detection~\citep{mastropaolo2021t5}.
CoTexT~\citep{phan2021cotext}, PLBART~\citep{ahmad2021plbart} based on BART, and CodeT5~\citep{wang2021codet5} based on T5 are representatives of encoder-decoder models for code.
AlphaCode~\citep{li2022alphacode} focused on solving competitive programming problems by adopting an encoder-decoder Transformer architecture to tackle code generation as a seq2seq translation task from a problem description to a solution source code because bidirectional representation can be helpful in understanding the problem description.
TransCoder~\citep{roziere2020transcoder} has demonstrated high performance in code translation by applying the back-translation technique used in machine translation~\citep{sennrich2016backtranslation}.

\paragraph{Other Models}

In addition to simply using pre-trained LMs, various techniques are applied to enhance code generation.
\citet{parvez2021retrieval} attempted to improve code generation and summarization using natural language retrieval techniques, and \citet{washio2022docs} and \citet{zhou2023docprompting} improved code generation for unknown libraries by retrieving the relevant documentation inspired by the behavior of human programmers referring to the API documentation while writing code.
Recent studies have incorporated reinforcement learning techniques to reward compilability of generated programs to ensure syntactic correctness~\citep{le2022coderl, wang2022compilable}.
Executing example test cases is a way to select a correct program from generated samples~\citep{li2022alphacode, shi2022execution}. TransCoder-ST~\citep{roziere2022transcoder-st} applied an automated unit test generation tool to filter out the invalid generated programs, and CodeT~\citep{chen2022codet} used an LLM to generate test cases along with programs to select the most suitable generated program.
UniXcoder~\citep{guo2022unixcoder} can change its architecture by switching between three modes, encoder-only, decoder-only, and encoder-decoder, to efficiently support various tasks.

\subsection{Solving Programming Problems Using LMs for Code}

LLMs pre-trained on a large amount of source code can perform several code-related tasks despite not being fine-tuned for specific tasks.
Solving programming problems can be treated as a text-to-code generation task, converting a problem description written in natural language into a solution program written in a programming language.

\citet{balog2017deepcoder} conducted early research on solving competitive programming problems using deep learning.
They proposed a framework, DeepCoder, that generates a program written in a domain-specific language from a problem description and several input/output examples.
AlphaCode~\citep{li2022alphacode} is a model focused on solving competitive programming problems, trained on GitHub source code and fine-tuned on a competitive programming dataset, CodeContests~\citep{li2022alphacode}, which includes CodeNet~\citep{ruchir2021codenet}.
It achieved a middle or above-middle ranking on average in programming competitions, and it is equivalent to an intermediate-level programmer.
Codex outscored most students~\citep{finnie2022robots} in solving introductory programming problems for first-year university students.

\subsection{Code Understanding of LMs for Code}
\label{sec:related-work:understanding}

The understanding of code by LMs is an important factor in their reliability across a range of applications, their generation results, and their robustness to prompts for practical use.
There are studies to investigate what information LMs retain, pay attention, or depend on.

\paragraph{Probing}

Probing is a diagnostic task aiming to investigate whether a pre-trained model contains specific information~\citep{karmakar2021knowledge}.
\citet{troshin2022probing} probed the code understanding of several widely used pre-trained models for code.
The results showed that they contain information about code syntactic structure and correctness, variable naming, etc.

\paragraph{Improvement Techniques}

\citet{kuznia2022lessismore} summarized problem descriptions to improve the quality of generated programs.
Codex outperformed the baseline (original problem descriptions) by 8.13\% by removing information such as human characters and background stories from the problem description.
However, the results indicated that the generation is affected by such information.

\paragraph{Memorization Issues}

There is concern about copying or memorization issues of the texts generated using GPT-based models~\citep{carlini2021extract,elangovan2021memorization}, as gaining interest in, such as leaking personal information~\citep{huang2022leaking} and license infringement~\citep{ciniselli2022copy}.

From the perspective of the infringement of open-source software licenses, \citet{ciniselli2022copy} reported that longer generated code is unlikely to be copied from the training data although this depends on the length of the generated code.
Similarly, in the NLP field, \citet{mccoy2021copy} showed that text generated using smaller n-gram models is substantially less novel than human-generated text; however, text generated using larger n-gram models is as novel as or even more novel than human-generated text.

However, as the most related work, \citet{karmakar2022hackerrank} stated that Codex could not solve the modified version of problems that was initially solved perfectly.
They also reported that some generated programs solve problems with completely different specifications, which is similar to our report on the inverse problem specification in Section~\ref{sec:results:modification:inversion}.
For example, although they modified the problem statement from \textit{"print the sum of the elements in a set A"} to \textit{"print the product of the elements in set A"}, the model still generated the solution for the original problem statement, i.e., generated a program to print the sum.
They stated that the Codex showed clear signs of memorization because it can also generate complete and correct programs even if significant information is missing.
Also, \citet{sontakke2022code} found that Transformer-based LMs, which aim to summarize source code, rely heavily on comments, function names, and variable names, and masking such information negatively impacts the generated summaries.

To defend against the memorization issue, \citet{lai2022ds1000} performed two perturbations (i.e., variations) of the collected problems as pre-processing to construct a benchmark dataset: surface perturbations (i.e., superficial modifications) and semantic perturbations (e.g., inverse problem specification).

\paragraph{Robustness}

ReCode~\citep{wang2022recode} is a benchmark dataset for evaluating robustness of code generation LMs.
Various perturbations (variations) were performed on the prompts given to the LMs to evaluate the robustness of the LMs to superficial modifications.
\citet{mouselinos2022bias} proposed a new framework to identify biases (i.e., defects of robustness) in several LLMs against several modifications for each block of a function: name block, description block, and example block.

\citet{mastropaolo2023robustness} investigated code generation in GitHub Copilot using \textit{different but semantically equivalent} (i.e., superficially modified) natural language descriptions.
They observed that the modifications impacted the correctness of the generated code by $\pm 28\%$.
Similarly, in arithmetic reasoning, \citet{shi2023distract} observed that the performance of LLMs is dramatically decreased by adding irrelevant sentences to the problem descriptions.

\section{Conclusion}
\label{sec:conclusion}

In this paper, we evaluated the robustness of popular LLMs for source code, Codex, CodeGen, InstructGPT, and ChatGPT, in solving programming problems by varying the prompt formats and modifying the problem descriptions. 

Our results showed that the newer models which incorporated the RLHF technique have stronger robustness to the formatting of problem descriptions.
On the other hand, Codex showed a 9.0\% performance improvement, representing the difference between the worst and best formats.
The prompt formatting, conducted on Codex and CodeGen, significantly affected the performance in both models, with a 25.9\% improvement in Codex and a 14.7\% improvement in CodeGen.

We also found that Codex heavily relies on the information of the variable names defined in the problems, as using unique variable names such as UUID variables caused the largest drop in the average solved rate, and the shuffled variable names had a similar impact.
This suggests that variable names play an important role in code generation performance.
Conversely, it is effective to use appropriate variable names in prompts, and we suggest not to use typical variable names in a wrong context.
Furthermore, we observed that the model rejected using the anonymized variable names in their generated programs while using the original variable names.

The results indicated that the solved rate is greatly affected by superficial modifications of problem descriptions in earlier models, which suggests that the models may not have a deep understanding of the problem description.
Nevertheless, we observed that Codex could adapt its solution approach in response to semantic modifications such as inverse problem specification.
However, we also observed that the model rarely generates programs solving more general problems that are the exact opposite of the given problem, which may be due to statistical biases in the training data.

Although an LLM can solve certain problems at a certain level, this does not necessarily imply that the model understands the problem and can solve other problems at the same level.
Our findings indicate that slight modifications to problem descriptions can significantly impact code generation performance; nonetheless, the newer models are gaining robustness, and the influence is being mitigated models evolve.

\section*{Acknowledgments}
\label{sec:acknowledgments}

This work was supported by the Japan Society for the Promotion of Science (JSPS) KAKENHI Grant Number JP23H03508.

\bibliography{main}{}

\begin{thebibliography}{81}
\providecommand{\natexlab}[1]{#1}
\providecommand{\url}[1]{\texttt{#1}}
\expandafter\ifx\csname urlstyle\endcsname\relax
  \providecommand{\doi}[1]{doi: #1}\else
  \providecommand{\doi}{doi: \begingroup \urlstyle{rm}\Url}\fi

\bibitem[Shirafuji et~al.(2022)Shirafuji, Ito, Morishita, Nakamura, Oda,
  Suzuki, and Watanobe]{shirafuji2022prompt}
Atsushi Shirafuji, Takumi Ito, Makoto Morishita, Yuki Nakamura, Yusuke Oda, Jun
  Suzuki, and Yutaka Watanobe.
\newblock Prompt sensitivity of language model for solving programming
  problems.
\newblock In \emph{Proceedings of the 21st International Conference on
  Intelligent Software Methodologies, Tools and Techniques (SOMET)}, pages
  346--359, 2022.

\bibitem[Chen et~al.(2021)Chen, Tworek, Jun, Yuan, de~Oliveira~Pinto, Kaplan,
  Edwards, Burda, Joseph, Brockman, Ray, Puri, Krueger, Petrov, Khlaaf, Sastry,
  Mishkin, Chan, Gray, Ryder, Pavlov, Power, Kaiser, Bavarian, Winter, Tillet,
  Such, Cummings, Plappert, Chantzis, Barnes, Herbert-Voss, Guss, Nichol,
  Paino, Tezak, Tang, Babuschkin, Balaji, Jain, Saunders, Hesse, Carr, Leike,
  Achiam, Misra, Morikawa, Radford, Knight, Brundage, Murati, Mayer, Welinder,
  McGrew, Amodei, McCandlish, Sutskever, and Zaremba]{chen2021codex}
Mark Chen, Jerry Tworek, Heewoo Jun, Qiming Yuan, Henrique~Ponde
  de~Oliveira~Pinto, Jared Kaplan, Harri Edwards, Yuri Burda, Nicholas Joseph,
  Greg Brockman, Alex Ray, Raul Puri, Gretchen Krueger, Michael Petrov, Heidy
  Khlaaf, Girish Sastry, Pamela Mishkin, Brooke Chan, Scott Gray, Nick Ryder,
  Mikhail Pavlov, Alethea Power, Lukasz Kaiser, Mohammad Bavarian, Clemens
  Winter, Philippe Tillet, Felipe~Petroski Such, Dave Cummings, Matthias
  Plappert, Fotios Chantzis, Elizabeth Barnes, Ariel Herbert-Voss,
  William~Hebgen Guss, Alex Nichol, Alex Paino, Nikolas Tezak, Jie Tang, Igor
  Babuschkin, Suchir Balaji, Shantanu Jain, William Saunders, Christopher
  Hesse, Andrew~N. Carr, Jan Leike, Josh Achiam, Vedant Misra, Evan Morikawa,
  Alec Radford, Matthew Knight, Miles Brundage, Mira Murati, Katie Mayer, Peter
  Welinder, Bob McGrew, Dario Amodei, Sam McCandlish, Ilya Sutskever, and
  Wojciech Zaremba.
\newblock Evaluating large language models trained on code.
\newblock \emph{arXiv preprint}, arXiv:2107.03374, 2021.

\bibitem[Parvez et~al.(2021)Parvez, Ahmad, Chakraborty, Ray, and
  Chang]{parvez2021retrieval}
Md~Rizwan Parvez, Wasi Ahmad, Saikat Chakraborty, Baishakhi Ray, and Kai-Wei
  Chang.
\newblock Retrieval augmented code generation and summarization.
\newblock In \emph{Findings of the Association for Computational Linguistics:
  EMNLP}, pages 2719--2734, 2021.

\bibitem[Nijkamp et~al.(2022)Nijkamp, Pang, Hayashi, Tu, Wang, Zhou, Savarese,
  and Xiong]{nijkamp2022codegen}
Erik Nijkamp, Bo~Pang, Hiroaki Hayashi, Lifu Tu, Huan Wang, Yingbo Zhou, Silvio
  Savarese, and Caiming Xiong.
\newblock {CodeGen}: An open large language model for code with multi-turn
  program synthesis.
\newblock \emph{arXiv preprint}, arXiv:2203.13474, 2022.

\bibitem[Fried et~al.(2023)Fried, Aghajanyan, Lin, Wang, Wallace, Shi, Zhong,
  Yih, Zettlemoyer, and Lewis]{fried2022incoder}
Daniel Fried, Armen Aghajanyan, Jessy Lin, Sida Wang, Eric Wallace, Freda Shi,
  Ruiqi Zhong, Scott Yih, Luke Zettlemoyer, and Mike Lewis.
\newblock {InCoder}: A generative model for code infilling and synthesis.
\newblock In \emph{International Conference on Learning Representations
  (ICLR)}, 2023.

\bibitem[Li et~al.(2022)Li, Choi, Chung, Kushman, Schrittwieser, Leblond,
  Eccles, Keeling, Gimeno, Lago, Hubert, Choy, de~Masson~d’Autume,
  Babuschkin, Chen, Huang, Welbl, Gowal, Cherepanov, Molloy, Mankowitz, Robson,
  Kohli, de~Freitas, Kavukcuoglu, and Vinyals]{li2022alphacode}
Yujia Li, David Choi, Junyoung Chung, Nate Kushman, Julian Schrittwieser, Rémi
  Leblond, Tom Eccles, James Keeling, Felix Gimeno, Agustin~Dal Lago, Thomas
  Hubert, Peter Choy, Cyprien de~Masson~d’Autume, Igor Babuschkin, Xinyun
  Chen, Po-Sen Huang, Johannes Welbl, Sven Gowal, Alexey Cherepanov, James
  Molloy, Daniel~J. Mankowitz, Esme~Sutherland Robson, Pushmeet Kohli, Nando
  de~Freitas, Koray Kavukcuoglu, and Oriol Vinyals.
\newblock Competition-level code generation with {AlphaCode}.
\newblock \emph{Science}, 378\penalty0 (6624):\penalty0 1092--1097, 2022.

\bibitem[Chen et~al.(2023)Chen, Zhang, Nguyen, Zan, Lin, Lou, and
  Chen]{chen2022codet}
Bei Chen, Fengji Zhang, Anh Nguyen, Daoguang Zan, Zeqi Lin, Jian-Guang Lou, and
  Weizhu Chen.
\newblock {CodeT}: Code generation with generated tests.
\newblock In \emph{International Conference on Learning Representations
  (ICLR)}, 2023.

\bibitem[Chowdhery et~al.(2022)Chowdhery, Narang, Devlin, Bosma, Mishra,
  Roberts, Barham, Chung, Sutton, Gehrmann, Schuh, Shi, Tsvyashchenko, Maynez,
  Rao, Barnes, Tay, Shazeer, Prabhakaran, Reif, Du, Hutchinson, Pope, Bradbury,
  Austin, Isard, Gur-Ari, Yin, Duke, Levskaya, Ghemawat, Dev, Michalewski,
  Garcia, Misra, Robinson, Fedus, Zhou, Ippolito, Luan, Lim, Zoph, Spiridonov,
  Sepassi, Dohan, Agrawal, Omernick, Dai, Pillai, Pellat, Lewkowycz, Moreira,
  Child, Polozov, Lee, Zhou, Wang, Saeta, Diaz, Firat, Catasta, Wei,
  Meier-Hellstern, Eck, Dean, Petrov, and Fiedel]{chowdhery2022palm}
Aakanksha Chowdhery, Sharan Narang, Jacob Devlin, Maarten Bosma, Gaurav Mishra,
  Adam Roberts, Paul Barham, Hyung~Won Chung, Charles Sutton, Sebastian
  Gehrmann, Parker Schuh, Kensen Shi, Sasha Tsvyashchenko, Joshua Maynez,
  Abhishek Rao, Parker Barnes, Yi~Tay, Noam Shazeer, Vinodkumar Prabhakaran,
  Emily Reif, Nan Du, Ben Hutchinson, Reiner Pope, James Bradbury, Jacob
  Austin, Michael Isard, Guy Gur-Ari, Pengcheng Yin, Toju Duke, Anselm
  Levskaya, Sanjay Ghemawat, Sunipa Dev, Henryk Michalewski, Xavier Garcia,
  Vedant Misra, Kevin Robinson, Liam Fedus, Denny Zhou, Daphne Ippolito, David
  Luan, Hyeontaek Lim, Barret Zoph, Alexander Spiridonov, Ryan Sepassi, David
  Dohan, Shivani Agrawal, Mark Omernick, Andrew~M. Dai,
  Thanumalayan~Sankaranarayana Pillai, Marie Pellat, Aitor Lewkowycz, Erica
  Moreira, Rewon Child, Oleksandr Polozov, Katherine Lee, Zongwei Zhou, Xuezhi
  Wang, Brennan Saeta, Mark Diaz, Orhan Firat, Michele Catasta, Jason Wei,
  Kathy Meier-Hellstern, Douglas Eck, Jeff Dean, Slav Petrov, and Noah Fiedel.
\newblock {PaLM}: Scaling language modeling with pathways.
\newblock \emph{arXiv preprint}, arXiv:2204.02311, 2022.

\bibitem[Christopoulou et~al.(2022)Christopoulou, Lampouras, Gritta, Zhang,
  Guo, Li, Zhang, Xiao, Shen, Li, Yu, Yan, Zhou, Wang, Ma, Iacobacci, Wang,
  Liang, Wei, Jiang, Wang, and Liu]{christopoulou2022pangucoder}
Fenia Christopoulou, Gerasimos Lampouras, Milan Gritta, Guchun Zhang, Yinpeng
  Guo, Zhongqi Li, Qi~Zhang, Meng Xiao, Bo~Shen, Lin Li, Hao Yu, Li~Yan, Pingyi
  Zhou, Xin Wang, Yuchi Ma, Ignacio Iacobacci, Yasheng Wang, Guangtai Liang,
  Jiansheng Wei, Xin Jiang, Qianxiang Wang, and Qun Liu.
\newblock {PanGu-Coder}: Program synthesis with function-level language
  modeling.
\newblock \emph{arXiv preprint}, arXiv:2207.11280, 2022.

\bibitem[Le et~al.(2022)Le, Wang, Gotmare, Savarese, and Hoi]{le2022coderl}
Hung Le, Yue Wang, Akhilesh~Deepak Gotmare, Silvio Savarese, and Steven
  Chu~Hong Hoi.
\newblock {CodeRL}: Mastering code generation through pretrained models and
  deep reinforcement learning.
\newblock In \emph{Advances in Neural Information Processing Systems
  (NeurIPS)}, volume~35, pages 21314--21328, 2022.

\bibitem[Xu et~al.(2022)Xu, Alon, Neubig, and Hellendoorn]{xu2022polycoder}
Frank~F. Xu, Uri Alon, Graham Neubig, and Vincent~Josua Hellendoorn.
\newblock A systematic evaluation of large language models of code.
\newblock In \emph{Deep Learning for Code Workshop}, 2022.

\bibitem[Wang et~al.(2021{\natexlab{a}})Wang, Wang, Joty, and
  Hoi]{wang2021codet5}
Yue Wang, Weishi Wang, Shafiq Joty, and Steven~C.H. Hoi.
\newblock {CodeT5}: Identifier-aware unified pre-trained encoder-decoder models
  for code understanding and generation.
\newblock In \emph{Proceedings of the Conference on Empirical Methods in
  Natural Language Processing (EMNLP)}, pages 8696--8708, 2021{\natexlab{a}}.

\bibitem[Wei et~al.(2019)Wei, Li, Xia, Fu, and Jin]{wei2019dual}
Bolin Wei, Ge~Li, Xin Xia, Zhiyi Fu, and Zhi Jin.
\newblock Code generation as a dual task of code summarization.
\newblock In \emph{Advances in Neural Information Processing Systems
  (NeurIPS)}, volume~32, 2019.

\bibitem[Lu et~al.(2021)Lu, Guo, Ren, Huang, Svyatkovskiy, Blanco, Clement,
  Drain, Jiang, Tang, Li, Zhou, Shou, Zhou, Tufano, GONG, Zhou, Duan,
  Sundaresan, Deng, Fu, and LIU]{lu2021codexglue}
Shuai Lu, Daya Guo, Shuo Ren, Junjie Huang, Alexey Svyatkovskiy, Ambrosio
  Blanco, Colin Clement, Dawn Drain, Daxin Jiang, Duyu Tang, Ge~Li, Lidong
  Zhou, Linjun Shou, Long Zhou, Michele Tufano, MING GONG, Ming Zhou, Nan Duan,
  Neel Sundaresan, Shao~Kun Deng, Shengyu Fu, and Shujie LIU.
\newblock {CodeXGLUE}: A machine learning benchmark dataset for code
  understanding and generation.
\newblock In \emph{Proceedings of the Neural Information Processing Systems
  Track on Datasets and Benchmarks}, volume~1, 2021.

\bibitem[Ahmad et~al.(2021)Ahmad, Chakraborty, Ray, and Chang]{ahmad2021plbart}
Wasi Ahmad, Saikat Chakraborty, Baishakhi Ray, and Kai-Wei Chang.
\newblock Unified pre-training for program understanding and generation.
\newblock In \emph{Proceedings of the Conference of the North American Chapter
  of the Association for Computational Linguistics: Human Language Technologies
  (NAACL-HLT)}, pages 2655--2668, 2021.

\bibitem[Roziere et~al.(2020)Roziere, Lachaux, Chanussot, and
  Lample]{roziere2020transcoder}
Baptiste Roziere, Marie-Anne Lachaux, Lowik Chanussot, and Guillaume Lample.
\newblock Unsupervised translation of programming languages.
\newblock In \emph{Advances in Neural Information Processing Systems
  (NeurIPS)}, volume~33, pages 20601--20611, 2020.

\bibitem[Roziere et~al.(2022)Roziere, Zhang, Charton, Harman, Synnaeve, and
  Lample]{roziere2022transcoder-st}
Baptiste Roziere, Jie Zhang, Francois Charton, Mark Harman, Gabriel Synnaeve,
  and Guillaume Lample.
\newblock Leveraging automated unit tests for unsupervised code translation.
\newblock In \emph{International Conference on Learning Representations
  (ICLR)}, 2022.

\bibitem[Svyatkovskiy et~al.(2020)Svyatkovskiy, Deng, Fu, and
  Sundaresan]{svyatkovskiy2020intellicode}
Alexey Svyatkovskiy, Shao~Kun Deng, Shengyu Fu, and Neel Sundaresan.
\newblock {IntelliCode Compose}: Code generation using transformer.
\newblock In \emph{Proceedings of the ACM Joint Meeting on European Software
  Engineering Conference and Symposium on the Foundations of Software
  Engineering (ESEC/FSE)}, pages 1433–--1443, 2020.

\bibitem[Terada and Watanobe(2021)]{terada2021completion}
Kenta Terada and Yutaka Watanobe.
\newblock Code completion for programming education based on deep learning.
\newblock \emph{International Journal of Computational Intelligence Studies},
  10\penalty0 (2-3):\penalty0 78--98, 2021.

\bibitem[Berabi et~al.(2021)Berabi, He, Raychev, and Vechev]{berabi2021tfix}
Berkay Berabi, Jingxuan He, Veselin Raychev, and Martin Vechev.
\newblock Tfix: Learning to fix coding errors with a text-to-text transformer.
\newblock In \emph{Proceedings of the 38th International Conference on Machine
  Learning (ICML)}, volume 139, pages 780--791, 2021.

\bibitem[Rahman et~al.(2021)Rahman, Watanobe, and Nakamura]{rahman2021repair}
Md.~Mostafizer Rahman, Yutaka Watanobe, and Keita Nakamura.
\newblock A bidirectional lstm language model for code evaluation and repair.
\newblock \emph{Symmetry}, 13\penalty0 (2), 2021.

\bibitem[Matsumoto et~al.(2021)Matsumoto, Watanobe, and
  Nakamura]{matsumoto2021repair}
Taku Matsumoto, Yutaka Watanobe, and Keita Nakamura.
\newblock A model with iterative trials for correcting logic errors in source
  code.
\newblock \emph{Applied Sciences}, 11\penalty0 (11), 2021.

\bibitem[Prenner et~al.(2022)Prenner, Babii, and Robbes]{prenner2021repair}
Julian~Aron Prenner, Hlib Babii, and Romain Robbes.
\newblock Can openai's codex fix bugs? an evaluation on quixbugs.
\newblock In \emph{Proceedings of the Third International Workshop on Automated
  Program Repair (APR)}, pages 69–--75, 2022.

\bibitem[Pearce et~al.(2023)Pearce, Tan, Ahmad, Karri, and
  Dolan-Gavitt]{pearce2021securitybugs}
H.~Pearce, B.~Tan, B.~Ahmad, R.~Karri, and B.~Dolan-Gavitt.
\newblock Examining zero-shot vulnerability repair with large language models.
\newblock In \emph{Proceedings of the 2023 IEEE Symposium on Security and
  Privacy (SP)}, pages 2339--2356, 2023.

\bibitem[Joshi et~al.(2022)Joshi, Cambronero, Gulwani, Le, Radicek, and
  Verbruggen]{joshi2022repair}
Harshit Joshi, José Cambronero, Sumit Gulwani, Vu~Le, Ivan Radicek, and Gust
  Verbruggen.
\newblock Repair is nearly generation: Multilingual program repair with llms.
\newblock \emph{arXiv preprint}, arXiv:2208.11640, 2022.

\bibitem[Shalaby et~al.(2017)Shalaby, Mehrez, El~Mougy, Abdulnasser, and
  Al-Safty]{shalaby2017algorithm}
Maged Shalaby, Tarek Mehrez, Amr El~Mougy, Khalid Abdulnasser, and Aysha
  Al-Safty.
\newblock Automatic algorithm recognition of source-code using machine
  learning.
\newblock In \emph{Proceedings of the 2017 16th IEEE International Conference
  on Machine Learning and Applications (ICMLA)}, pages 170--177, 2017.

\bibitem[Chourasia et~al.(2022)Chourasia, Ramakrishnan, Apte, and
  Kumar]{chourasia2022algorithm}
Pranshu Chourasia, Ganesh Ramakrishnan, Varsha Apte, and Suraj Kumar.
\newblock Algorithm identification in programming assignments.
\newblock In \emph{Proceedings of the 30th IEEE/ACM International Conference on
  Program Comprehension (ICPC)}, page 471–481, 2022.

\bibitem[Watanobe et~al.(2023)Watanobe, Rahman, Amin, and
  Kabir]{watanobe2023algorithm}
Yutaka Watanobe, Md~Mostafizer Rahman, Md~Faizul~Ibne Amin, and Raihan Kabir.
\newblock Identifying algorithm in program code based on structural features
  using cnn classification model.
\newblock \emph{Applied Intelligence}, 53\penalty0 (10):\penalty0 12210--12236,
  2023.

\bibitem[Huang et~al.(2022)Huang, Shao, and Chang]{huang2022leaking}
Jie Huang, Hanyin Shao, and Kevin Chen-Chuan Chang.
\newblock Are large pre-trained language models leaking your personal
  information?
\newblock In \emph{Findings of the Association for Computational Linguistics:
  EMNLP}, pages 2038--2047, 2022.

\bibitem[Ciniselli et~al.(2022)Ciniselli, Pascarella, and
  Bavota]{ciniselli2022copy}
Matteo Ciniselli, Luca Pascarella, and Gabriele Bavota.
\newblock To what extent do deep learning-based code recommenders generate
  predictions by cloning code from the training set?
\newblock In \emph{Proceedings of the 19th International Conference on Mining
  Software Repositories (MSR)}, pages 167–--178, 2022.

\bibitem[Carlini et~al.(2021)Carlini, Tram{\`e}r, Wallace, Jagielski,
  Herbert-Voss, Lee, Roberts, Brown, Song, Erlingsson, Oprea, and
  Raffel]{carlini2021extract}
Nicholas Carlini, Florian Tram{\`e}r, Eric Wallace, Matthew Jagielski, Ariel
  Herbert-Voss, Katherine Lee, Adam Roberts, Tom Brown, Dawn Song, {\'U}lfar
  Erlingsson, Alina Oprea, and Colin Raffel.
\newblock Extracting training data from large language models.
\newblock In \emph{30th USENIX Security Symposium (USENIX Security)}, pages
  2633--2650, 2021.

\bibitem[Elangovan et~al.(2021)Elangovan, He, and
  Verspoor]{elangovan2021memorization}
Aparna Elangovan, Jiayuan He, and Karin Verspoor.
\newblock Memorization vs. generalization : Quantifying data leakage in {NLP}
  performance evaluation.
\newblock In \emph{Proceedings of the 16th Conference of the European Chapter
  of the Association for Computational Linguistics: Main Volume}, pages
  1325--1335, 2021.

\bibitem[Karmakar et~al.(2022)Karmakar, Prenner, D'Ambros, and
  Robbes]{karmakar2022hackerrank}
Anjan Karmakar, Julian~Aron Prenner, Marco D'Ambros, and Romain Robbes.
\newblock Codex hacks hackerrank: Memorization issues and a framework for code
  synthesis evaluation.
\newblock \emph{arXiv preprint}, arXiv:2212.02684, 2022.

\bibitem[Mastropaolo et~al.(2023)Mastropaolo, Pascarella, Guglielmi, Ciniselli,
  Scalabrino, Oliveto, and Bavota]{mastropaolo2023robustness}
Antonio Mastropaolo, Luca Pascarella, Emanuela Guglielmi, Matteo Ciniselli,
  Simone Scalabrino, Rocco Oliveto, and Gabriele Bavota.
\newblock On the robustness of code generation techniques: An empirical study
  on github copilot.
\newblock \emph{arXiv preprint}, arXiv:2302.00438, 2023.

\bibitem[Wang et~al.(2022{\natexlab{a}})Wang, Li, Qian, Yang, Wang, Shang,
  Kumar, Tan, Ray, Bhatia, Nallapati, Ramanathan, Roth, and
  Xiang]{wang2022recode}
Shiqi Wang, Zheng Li, Haifeng Qian, Chenghao Yang, Zijian Wang, Mingyue Shang,
  Varun Kumar, Samson Tan, Baishakhi Ray, Parminder Bhatia, Ramesh Nallapati,
  Murali~Krishna Ramanathan, Dan Roth, and Bing Xiang.
\newblock {ReCode}: Robustness evaluation of code generation models.
\newblock \emph{arXiv preprint}, arXiv:2212.10264, 2022{\natexlab{a}}.

\bibitem[Ouyang et~al.(2022)Ouyang, Wu, Jiang, Almeida, Wainwright, Mishkin,
  Zhang, Agarwal, Slama, Ray, Schulman, Hilton, Kelton, Miller, Simens, Askell,
  Welinder, Christiano, Leike, and Lowe]{ouyang2022instructgpt}
Long Ouyang, Jeffrey Wu, Xu~Jiang, Diogo Almeida, Carroll Wainwright, Pamela
  Mishkin, Chong Zhang, Sandhini Agarwal, Katarina Slama, Alex Ray, John
  Schulman, Jacob Hilton, Fraser Kelton, Luke Miller, Maddie Simens, Amanda
  Askell, Peter Welinder, Paul~F Christiano, Jan Leike, and Ryan Lowe.
\newblock Training language models to follow instructions with human feedback.
\newblock In \emph{Advances in Neural Information Processing Systems
  (NeurIPS)}, volume~35, pages 27730--27744, 2022.

\bibitem[Watanobe(2004)]{watanobe2004aoj}
Yutaka Watanobe.
\newblock {Aizu} {Online} {Judge}, 2004.
\newblock URL \url{https://onlinejudge.u-aizu.ac.jp/}.

\bibitem[Watanobe et~al.(2022)Watanobe, Rahman, Matsumoto, Rage, and
  Ravikumar]{watanobe2022aoj}
Yutaka Watanobe, Md.~Mostafizer Rahman, Taku Matsumoto, Uday~Kiran Rage, and
  Penugonda Ravikumar.
\newblock Online judge system: Requirements, architecture, and experiences.
\newblock \emph{International Journal of Software Engineering and Knowledge
  Engineering}, 32\penalty0 (4):\penalty0 1--30, 2022.

\bibitem[Wasik et~al.(2018)Wasik, Antczak, Badura, Laskowski, and
  Sternal]{wasik2018oj}
Szymon Wasik, Maciej Antczak, Jan Badura, Artur Laskowski, and Tomasz Sternal.
\newblock A survey on online judge systems and their applications.
\newblock \emph{ACM Comput. Surv.}, 51\penalty0 (1), 2018.

\bibitem[Vaswani et~al.(2017)Vaswani, Shazeer, Parmar, Uszkoreit, Jones, Gomez,
  Kaiser, and Polosukhin]{vaswani2017transformer}
Ashish Vaswani, Noam Shazeer, Niki Parmar, Jakob Uszkoreit, Llion Jones,
  Aidan~N Gomez, \L~ukasz Kaiser, and Illia Polosukhin.
\newblock Attention is all you need.
\newblock In \emph{Advances in Neural Information Processing Systems (NIPS)},
  volume~30, 2017.

\bibitem[OpenAI(2023)]{openai2023gpt-4}
OpenAI.
\newblock {GPT-4} technical report.
\newblock \emph{arXiv preprint}, arXiv:2303.08774, 2023.

\bibitem[Brown et~al.(2020)Brown, Mann, Ryder, Subbiah, Kaplan, Dhariwal,
  Neelakantan, Shyam, Sastry, Askell, Agarwal, Herbert-Voss, Krueger, Henighan,
  Child, Ramesh, Ziegler, Wu, Winter, Hesse, Chen, Sigler, Litwin, Gray, Chess,
  Clark, Berner, McCandlish, Radford, Sutskever, and Amodei]{brown2020gpt3}
Tom Brown, Benjamin Mann, Nick Ryder, Melanie Subbiah, Jared~D Kaplan, Prafulla
  Dhariwal, Arvind Neelakantan, Pranav Shyam, Girish Sastry, Amanda Askell,
  Sandhini Agarwal, Ariel Herbert-Voss, Gretchen Krueger, Tom Henighan, Rewon
  Child, Aditya Ramesh, Daniel Ziegler, Jeffrey Wu, Clemens Winter, Chris
  Hesse, Mark Chen, Eric Sigler, Mateusz Litwin, Scott Gray, Benjamin Chess,
  Jack Clark, Christopher Berner, Sam McCandlish, Alec Radford, Ilya Sutskever,
  and Dario Amodei.
\newblock Language models are few-shot learners.
\newblock In \emph{Proceedings of Advances in Neural Information Processing
  Systems (NeurIPS)}, volume~33, pages 1877--1901, 2020.

\bibitem[Hendrycks et~al.(2021)Hendrycks, Basart, Kadavath, Mazeika, Arora,
  Guo, Burns, Puranik, He, Song, and Steinhardt]{hendrycks2021apps}
Dan Hendrycks, Steven Basart, Saurav Kadavath, Mantas Mazeika, Akul Arora,
  Ethan Guo, Collin Burns, Samir Puranik, Horace He, Dawn Song, and Jacob
  Steinhardt.
\newblock Measuring coding challenge competence with {APPS}.
\newblock In \emph{Proceedings of the Neural Information Processing Systems
  Track on Datasets and Benchmarks}, volume~1, 2021.

\bibitem[Drori and Verma(2021)]{drori2021algebra}
Iddo Drori and Nakul Verma.
\newblock Solving linear algebra by program synthesis.
\newblock \emph{arXiv preprint}, arXiv:2111.08171, 2021.

\bibitem[Tang et~al.(2022)Tang, Ke, Singh, Feng, Austin, Verma, and
  Drori]{tang2021probstat}
Leonard Tang, Elizabeth Ke, Nikhil Singh, Bo~Feng, Derek Austin, Nakul Verma,
  and Iddo Drori.
\newblock Solving probability and statistics problems by probabilistic program
  synthesis at human level and predicting solvability.
\newblock In \emph{Artificial Intelligence in Education. Posters and Late
  Breaking Results, Workshops and Tutorials, Industry and Innovation Tracks,
  Practitioners’ and Doctoral Consortium}, pages 612–--615, 2022.

\bibitem[Drori et~al.(2022)Drori, Zhang, Shuttleworth, Tang, Lu, Ke, Liu, Chen,
  Tran, Cheng, Wang, Singh, Patti, Lynch, Shporer, Verma, Wu, and
  Strang]{drori2021math}
Iddo Drori, Sarah Zhang, Reece Shuttleworth, Leonard Tang, Albert Lu, Elizabeth
  Ke, Kevin Liu, Linda Chen, Sunny Tran, Newman Cheng, Roman Wang, Nikhil
  Singh, Taylor~L. Patti, Jayson Lynch, Avi Shporer, Nakul Verma, Eugene Wu,
  and Gilbert Strang.
\newblock A neural network solves, explains, and generates university math
  problems by program synthesis and few-shot learning at human level.
\newblock \emph{Proceedings of the National Academy of Sciences of the United
  States of America}, 119\penalty0 (32):\penalty0 e2123433119, 2022.

\bibitem[Sarsa et~al.(2022)Sarsa, Denny, Hellas, and
  Leinonen]{sarsa2022exercise}
Sami Sarsa, Paul Denny, Arto Hellas, and Juho Leinonen.
\newblock Automatic generation of programming exercises and code explanations
  using large language models.
\newblock In \emph{Proceedings of the 2022 ACM Conference on International
  Computing Education Research (ICER) - Volume 1}, pages 27--–43, 2022.

\bibitem[Gao et~al.(2020)Gao, Biderman, Black, Golding, Hoppe, Foster, Phang,
  He, Thite, Nabeshima, Presser, and Leahy]{gao2021thepile}
Leo Gao, Stella Biderman, Sid Black, Laurence Golding, Travis Hoppe, Charles
  Foster, Jason Phang, Horace He, Anish Thite, Noa Nabeshima, Shawn Presser,
  and Connor Leahy.
\newblock {The Pile}: An 800gb dataset of diverse text for language modeling.
\newblock \emph{arXiv preprint}, arXiv:2101.00027, 2020.

\bibitem[Papineni et~al.(2002)Papineni, Roukos, Ward, and
  Zhu]{papineni2002bleu}
Kishore Papineni, Salim Roukos, Todd Ward, and Wei-Jing Zhu.
\newblock {BLEU}: A method for automatic evaluation of machine translation.
\newblock In \emph{Proceedings of the 40th Annual Meeting on Association for
  Computational Linguistics (ACL)}, pages 311--318, 2002.

\bibitem[Ren et~al.(2020)Ren, Guo, Lu, Zhou, Liu, Tang, Sundaresan, Zhou,
  Blanco, and Ma]{ren2020codebleu}
Shuo Ren, Daya Guo, Shuai Lu, Long Zhou, Shujie Liu, Duyu Tang, Neel
  Sundaresan, Ming Zhou, Ambrosio Blanco, and Shuai Ma.
\newblock {CodeBLEU}: A method for automatic evaluation of code synthesis.
\newblock \emph{arXiv preprint}, arXiv:2009.10297, 2020.

\bibitem[Kulal et~al.(2019)Kulal, Pasupat, Chandra, Lee, Padon, Aiken, and
  Liang]{kulal2019spoc}
Sumith Kulal, Panupong Pasupat, Kartik Chandra, Mina Lee, Oded Padon, Alex
  Aiken, and Percy~S Liang.
\newblock Spoc: Search-based pseudocode to code.
\newblock In \emph{Advances in Neural Information Processing Systems},
  volume~32, 2019.

\bibitem[Pearce et~al.(2022)Pearce, Ahmad, Tan, Dolan-Gavitt, and
  Karri]{pearce2021security}
H.~Pearce, B.~Ahmad, B.~Tan, B.~Dolan-Gavitt, and R.~Karri.
\newblock Asleep at the keyboard? assessing the security of github copilot's
  code contributions.
\newblock In \emph{Proceedings of the IEEE Symposium on Security and Privacy
  (SP)}, pages 980--994, 2022.

\bibitem[Leach et~al.(2005)Leach, Salz, and Mealling]{leach2005uuid}
Paul~J. Leach, Rich Salz, and Michael~H. Mealling.
\newblock {A Universally Unique IDentifier (UUID) URN Namespace}.
\newblock RFC 4122, 2005.

\bibitem[Troshin and Chirkova(2022)]{troshin2022probing}
Sergey Troshin and Nadezhda Chirkova.
\newblock Probing pretrained models of source codes.
\newblock In \emph{Proceedings of the Fifth BlackboxNLP Workshop on Analyzing
  and Interpreting Neural Networks for NLP}, pages 371--383, 2022.

\bibitem[Devlin et~al.(2019)Devlin, Chang, Lee, and Toutanova]{devlin2019bert}
Jacob Devlin, Ming-Wei Chang, Kenton Lee, and Kristina Toutanova.
\newblock {BERT}: Pre-training of deep bidirectional transformers for language
  understanding.
\newblock In \emph{Proceedings of the 2019 Conference of the North {A}merican
  Chapter of the Association for Computational Linguistics: Human Language
  Technologies, Volume 1 (Long and Short Papers)}, pages 4171--4186, 2019.

\bibitem[Liu et~al.(2019)Liu, Ott, Goyal, Du, Joshi, Chen, Levy, Lewis,
  Zettlemoyer, and Stoyanov]{liu2019roberta}
Yinhan Liu, Myle Ott, Naman Goyal, Jingfei Du, Mandar Joshi, Danqi Chen, Omer
  Levy, Mike Lewis, Luke Zettlemoyer, and Veselin Stoyanov.
\newblock {RoBERTa}: A robustly optimized {BERT} pretraining approach.
\newblock \emph{arXiv preprint}, arXiv:1907.11692, 2019.

\bibitem[Feng et~al.(2020)Feng, Guo, Tang, Duan, Feng, Gong, Shou, Qin, Liu,
  Jiang, and Zhou]{feng2020codebert}
Zhangyin Feng, Daya Guo, Duyu Tang, Nan Duan, Xiaocheng Feng, Ming Gong, Linjun
  Shou, Bing Qin, Ting Liu, Daxin Jiang, and Ming Zhou.
\newblock {C}ode{BERT}: A pre-trained model for programming and natural
  languages.
\newblock In \emph{Findings of the Association for Computational Linguistics:
  EMNLP}, pages 1536--1547, 2020.

\bibitem[Kanade et~al.(2020)Kanade, Maniatis, Balakrishnan, and
  Shi]{kanade2020cubert}
Aditya Kanade, Petros Maniatis, Gogul Balakrishnan, and Kensen Shi.
\newblock Learning and evaluating contextual embedding of source code.
\newblock In \emph{Proceedings of the 37th International Conference on Machine
  Learning (ICML)}, volume 119, pages 5110--5121, 2020.

\bibitem[Wang et~al.(2021{\natexlab{b}})Wang, Wang, Mi, Zhou, Wan, Liu, Li, Wu,
  Liu, and Jiang]{wang2021syncobert}
Xin Wang, Yasheng Wang, Fei Mi, Pingyi Zhou, Yao Wan, Xiao Liu, Li~Li, Hao Wu,
  Jin Liu, and Xin Jiang.
\newblock {SynCoBERT}: Syntax-guided multi-modal contrastive pre-training for
  code representation.
\newblock \emph{arXiv preprint}, arXiv:2108.04556, 2021{\natexlab{b}}.

\bibitem[Jiang et~al.(2021)Jiang, Zheng, Lyu, Li, and Lyu]{jiang2021treebert}
Xue Jiang, Zhuoran Zheng, Chen Lyu, Liang Li, and Lei Lyu.
\newblock {TreeBERT}: A tree-based pre-trained model for programming language.
\newblock In \emph{Proceedings of the Thirty-Seventh Conference on Uncertainty
  in Artificial Intelligence}, volume 161, pages 54--63, 2021.

\bibitem[Guo et~al.(2021)Guo, Ren, Lu, Feng, Tang, LIU, Zhou, Duan,
  Svyatkovskiy, Fu, Tufano, Deng, Clement, Drain, Sundaresan, Yin, Jiang, and
  Zhou]{guo2021graphcodebert}
Daya Guo, Shuo Ren, Shuai Lu, Zhangyin Feng, Duyu Tang, Shujie LIU, Long Zhou,
  Nan Duan, Alexey Svyatkovskiy, Shengyu Fu, Michele Tufano, Shao~Kun Deng,
  Colin Clement, Dawn Drain, Neel Sundaresan, Jian Yin, Daxin Jiang, and Ming
  Zhou.
\newblock {GraphCodeBERT}: Pre-training code representations with data flow.
\newblock In \emph{International Conference on Learning Representations
  (ICLR)}, 2021.

\bibitem[Lewis et~al.(2020)Lewis, Liu, Goyal, Ghazvininejad, Mohamed, Levy,
  Stoyanov, and Zettlemoyer]{lewis2020bart}
Mike Lewis, Yinhan Liu, Naman Goyal, Marjan Ghazvininejad, Abdelrahman Mohamed,
  Omer Levy, Veselin Stoyanov, and Luke Zettlemoyer.
\newblock {BART}: Denoising sequence-to-sequence pre-training for natural
  language generation, translation, and comprehension.
\newblock In \emph{Proceedings of the 58th Annual Meeting of the Association
  for Computational Linguistics}, pages 7871--7880, 2020.

\bibitem[Raffel et~al.(2020)Raffel, Shazeer, Roberts, Lee, Narang, Matena,
  Zhou, Li, and Liu]{raffel2020t5}
Colin Raffel, Noam Shazeer, Adam Roberts, Katherine Lee, Sharan Narang, Michael
  Matena, Yanqi Zhou, Wei Li, and Peter~J. Liu.
\newblock Exploring the limits of transfer learning with a unified text-to-text
  transformer.
\newblock \emph{Journal of Machine Learning Research}, 21\penalty0
  (140):\penalty0 1--67, 2020.

\bibitem[Mastropaolo et~al.(2021)Mastropaolo, Scalabrino, Cooper,
  Nader~Palacio, Poshyvanyk, Oliveto, and Bavota]{mastropaolo2021t5}
Antonio Mastropaolo, Simone Scalabrino, Nathan Cooper, David Nader~Palacio,
  Denys Poshyvanyk, Rocco Oliveto, and Gabriele Bavota.
\newblock Studying the usage of text-to-text transfer transformer to support
  code-related tasks.
\newblock In \emph{2021 IEEE/ACM 43rd International Conference on Software
  Engineering (ICSE)}, pages 336--347, 2021.

\bibitem[Phan et~al.(2021)Phan, Tran, Le, Nguyen, Annibal, Peltekian, and
  Ye]{phan2021cotext}
Long Phan, Hieu Tran, Daniel Le, Hieu Nguyen, James Annibal, Alec Peltekian,
  and Yanfang Ye.
\newblock {CoTexT}: Multi-task learning with code-text transformer.
\newblock In \emph{Proceedings of the 1st Workshop on Natural Language
  Processing for Programming (NLP4Prog)}, pages 40--47, 2021.

\bibitem[Sennrich et~al.(2016)Sennrich, Haddow, and
  Birch]{sennrich2016backtranslation}
Rico Sennrich, Barry Haddow, and Alexandra Birch.
\newblock Improving neural machine translation models with monolingual data.
\newblock In \emph{Proceedings of the 54th Annual Meeting of the Association
  for Computational Linguistics (Volume 1: Long Papers)}, pages 86--96, 2016.

\bibitem[Washio and Miyao(2022)]{washio2022docs}
Koki Washio and Yusuke Miyao.
\newblock Code generation for unknown libraries via reading api documentations.
\newblock \emph{arXiv preprint}, arXiv:2202.07806, 2022.

\bibitem[Zhou et~al.(2023)Zhou, Alon, Xu, Jiang, and
  Neubig]{zhou2023docprompting}
Shuyan Zhou, Uri Alon, Frank~F. Xu, Zhengbao Jiang, and Graham Neubig.
\newblock {DocPrompting}: Generating code by retrieving the docs.
\newblock In \emph{International Conference on Learning Representations
  (ICLR)}, 2023.

\bibitem[Wang et~al.(2022{\natexlab{b}})Wang, Wang, Wan, Mi, Li, Zhou, Liu, Wu,
  Jiang, and Liu]{wang2022compilable}
Xin Wang, Yasheng Wang, Yao Wan, Fei Mi, Yitong Li, Pingyi Zhou, Jin Liu, Hao
  Wu, Xin Jiang, and Qun Liu.
\newblock Compilable neural code generation with compiler feedback.
\newblock In \emph{Findings of the Association for Computational Linguistics:
  ACL}, pages 9--19, 2022{\natexlab{b}}.

\bibitem[Shi et~al.(2022)Shi, Fried, Ghazvininejad, Zettlemoyer, and
  Wang]{shi2022execution}
Freda Shi, Daniel Fried, Marjan Ghazvininejad, Luke Zettlemoyer, and Sida~I.
  Wang.
\newblock Natural language to code translation with execution.
\newblock In \emph{Proceedings of the 2022 Conference on Empirical Methods in
  Natural Language Processing}, pages 3533--3546, 2022.

\bibitem[Guo et~al.(2022)Guo, Lu, Duan, Wang, Zhou, and Yin]{guo2022unixcoder}
Daya Guo, Shuai Lu, Nan Duan, Yanlin Wang, Ming Zhou, and Jian Yin.
\newblock {UniXcoder}: Unified cross-modal pre-training for code
  representation.
\newblock In \emph{Proceedings of the 60th Annual Meeting of the Association
  for Computational Linguistics (Volume 1: Long Papers)}, pages 7212--7225,
  2022.

\bibitem[Balog et~al.(2017)Balog, Gaunt, Brockschmidt, Nowozin, and
  Tarlow]{balog2017deepcoder}
Matej Balog, Alexander~L. Gaunt, Marc Brockschmidt, Sebastian Nowozin, and
  Daniel Tarlow.
\newblock {DeepCoder}: Learning to write programs.
\newblock In \emph{International Conference on Learning Representations
  (ICLR)}, 2017.

\bibitem[Puri et~al.(2021)Puri, Kung, Janssen, Zhang, Domeniconi, Zolotov,
  Dolby, Chen, Choudhury, Decker, Thost, Thost, Buratti, Pujar, Ramji, Finkler,
  Malaika, and Reiss]{ruchir2021codenet}
Ruchir Puri, David Kung, Geert Janssen, Wei Zhang, Giacomo Domeniconi, Vladimir
  Zolotov, Julian~T Dolby, Jie Chen, Mihir Choudhury, Lindsey Decker, Veronika
  Thost, Veronika Thost, Luca Buratti, Saurabh Pujar, Shyam Ramji, Ulrich
  Finkler, Susan Malaika, and Frederick Reiss.
\newblock {CodeNet}: A large-scale ai for code dataset for learning a diversity
  of coding tasks.
\newblock In \emph{Proceedings of the Neural Information Processing Systems
  Track on Datasets and Benchmarks}, volume~1, 2021.

\bibitem[Finnie-Ansley et~al.(2022)Finnie-Ansley, Denny, Becker, Luxton-Reilly,
  and Prather]{finnie2022robots}
James Finnie-Ansley, Paul Denny, Brett~A. Becker, Andrew Luxton-Reilly, and
  James Prather.
\newblock The robots are coming: Exploring the implications of openai codex on
  introductory programming.
\newblock In \emph{Proceedings of the Australasian Computing Education
  Conference (ACE)}, pages 10--–19, 2022.

\bibitem[Karmakar and Robbes(2021)]{karmakar2021knowledge}
Anjan Karmakar and Romain Robbes.
\newblock What do pre-trained code models know about code?
\newblock In \emph{2021 36th IEEE/ACM International Conference on Automated
  Software Engineering (ASE)}, pages 1332--1336, 2021.

\bibitem[Kuznia et~al.(2022)Kuznia, Mishra, Parmar, and
  Baral]{kuznia2022lessismore}
Kirby Kuznia, Swaroop Mishra, Mihir Parmar, and Chitta Baral.
\newblock Less is more: Summary of long instructions is better for program
  synthesis.
\newblock In \emph{Proceedings of the 2022 Conference on Empirical Methods in
  Natural Language Processing}, pages 4532--4552, 2022.

\bibitem[McCoy et~al.(2021)McCoy, Smolensky, Linzen, Gao, and
  Celikyilmaz]{mccoy2021copy}
R.~Thomas McCoy, Paul Smolensky, Tal Linzen, Jianfeng Gao, and Asli
  Celikyilmaz.
\newblock How much do language models copy from their training data? evaluating
  linguistic novelty in text generation using {RAVEN}.
\newblock \emph{arXiv preprint}, arXiv:2111.09509, 2021.

\bibitem[Sontakke et~al.(2022)Sontakke, Patwardhan, Vig, Medicherla, Naik, and
  Shroff]{sontakke2022code}
Ankita~Nandkishor Sontakke, Manasi Patwardhan, Lovekesh Vig, Raveendra~Kumar
  Medicherla, Ravindra Naik, and Gautam Shroff.
\newblock Code summarization: Do transformers really understand code?
\newblock In \emph{Deep Learning for Code Workshop}, 2022.

\bibitem[Lai et~al.(2022)Lai, Li, Wang, Zhang, Zhong, Zettlemoyer, tau Yih,
  Fried, Wang, and Yu]{lai2022ds1000}
Yuhang Lai, Chengxi Li, Yiming Wang, Tianyi Zhang, Ruiqi Zhong, Luke
  Zettlemoyer, Scott~Wen tau Yih, Daniel Fried, Sida Wang, and Tao Yu.
\newblock {DS-1000}: A natural and reliable benchmark for data science code
  generation.
\newblock \emph{arXiv preprint}, arXiv:2211.11501, 2022.

\bibitem[Mouselinos et~al.(2022)Mouselinos, Malinowski, and
  Michalewski]{mouselinos2022bias}
Spyridon Mouselinos, Mateusz Malinowski, and Henryk Michalewski.
\newblock A simple, yet effective approach to finding biases in code
  generation.
\newblock \emph{arXiv preprint}, arXiv:2211.00609, 2022.

\bibitem[Shi et~al.(2023)Shi, Chen, Misra, Scales, Dohan, Chi, Schärli, and
  Zhou]{shi2023distract}
Freda Shi, Xinyun Chen, Kanishka Misra, Nathan Scales, David Dohan, Ed~Chi,
  Nathanael Schärli, and Denny Zhou.
\newblock Large language models can be easily distracted by irrelevant context.
\newblock \emph{arXiv preprint}, arXiv:2302.00093, 2023.

\end{thebibliography}
\bibliographystyle{unsrtnat}

\newpage

\appendix

\section{Problem Formatting Examples}
\label{sec:appendix:formatting}

\begin{figure*}[ht]
    \centering
    \includegraphics[width=0.8\linewidth]{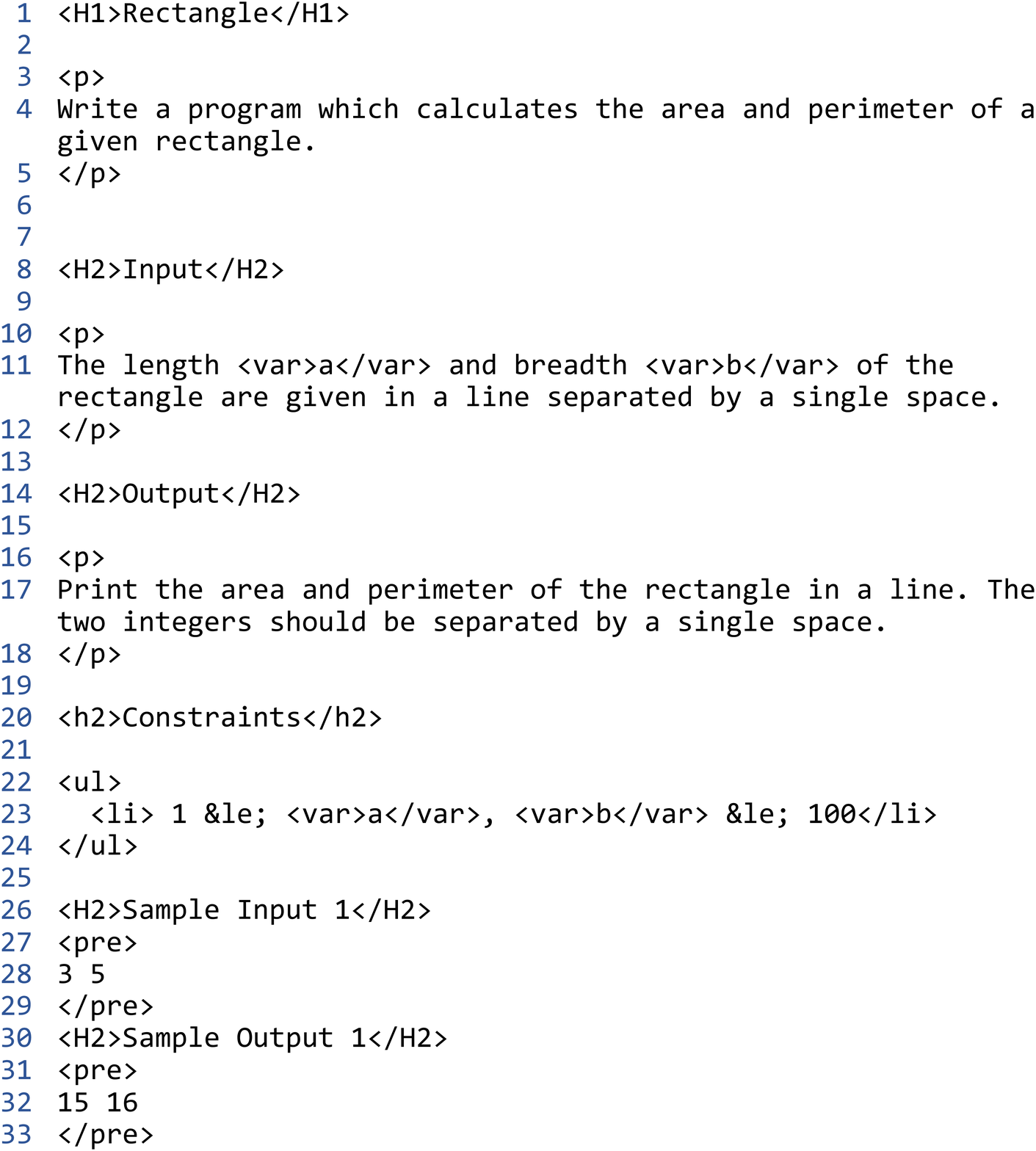}
    \caption{Example of Raw HTML problem (ITP1\_1\_C).}
    \label{fig:raw-problem}
\end{figure*}

\begin{figure*}[ht]
    \centering
    \includegraphics[width=0.8\linewidth]{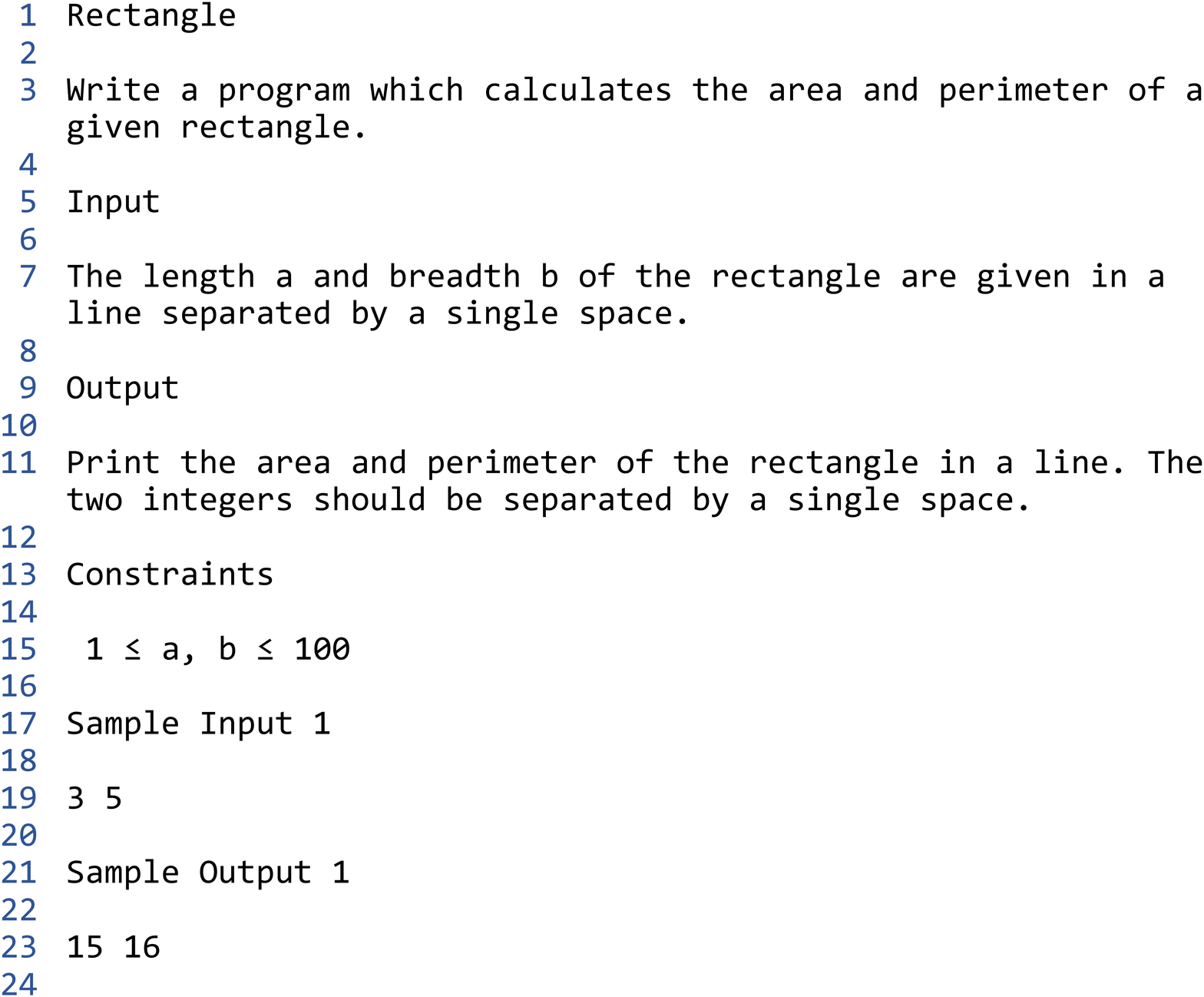}
    \caption{Example of parsed plain problem (ITP1\_1\_C).}
    \label{fig:parsed-problem}
\end{figure*}

\begin{figure*}[ht]
    \centering
    \includegraphics[width=0.8\linewidth]{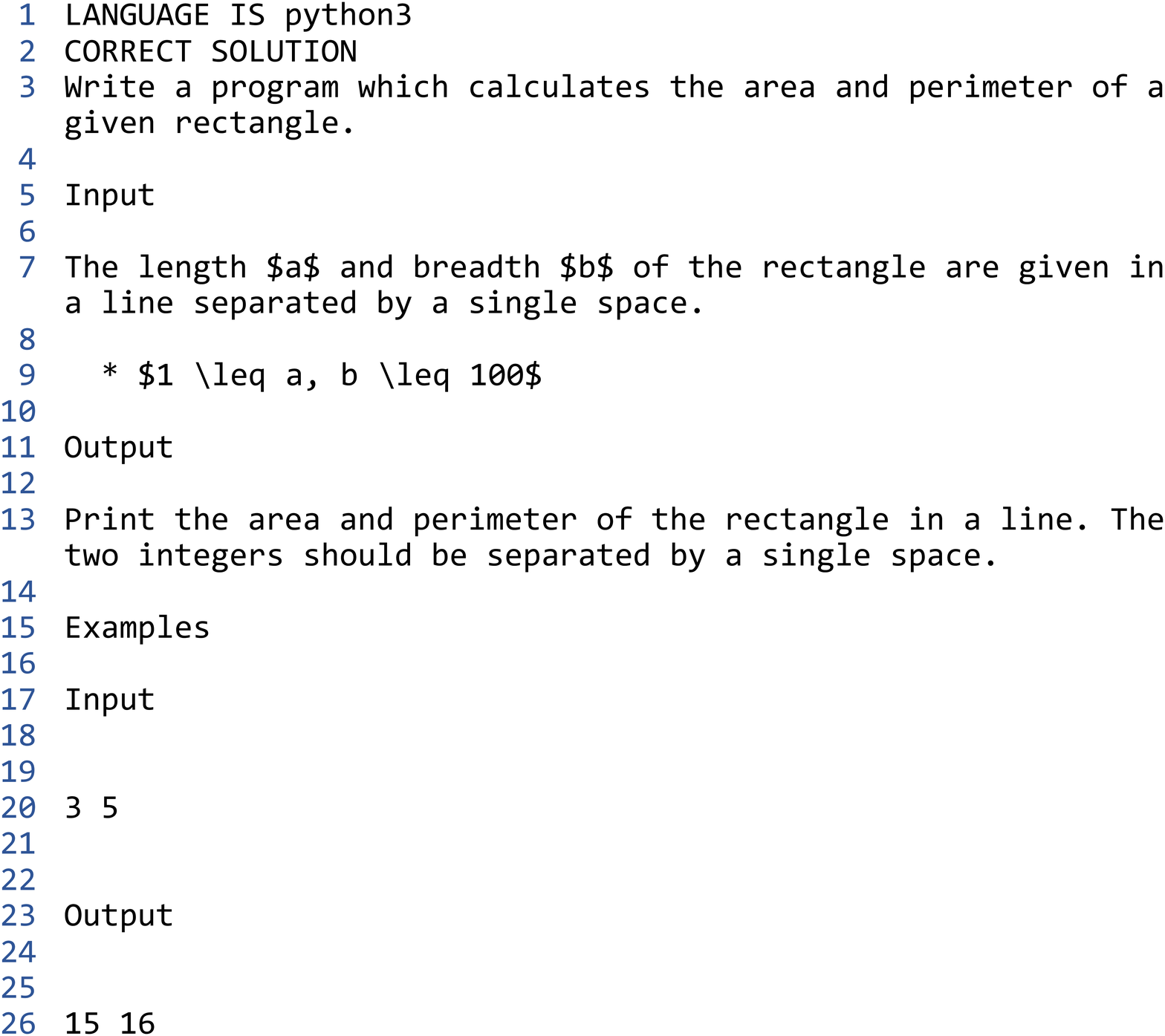}
    \caption{Example of AlphaCode-inspired problem (ITP1\_1\_C).}
    \label{fig:alphacode-problem}
\end{figure*}

\begin{figure*}[ht]
    \centering
    \includegraphics[width=0.8\linewidth]{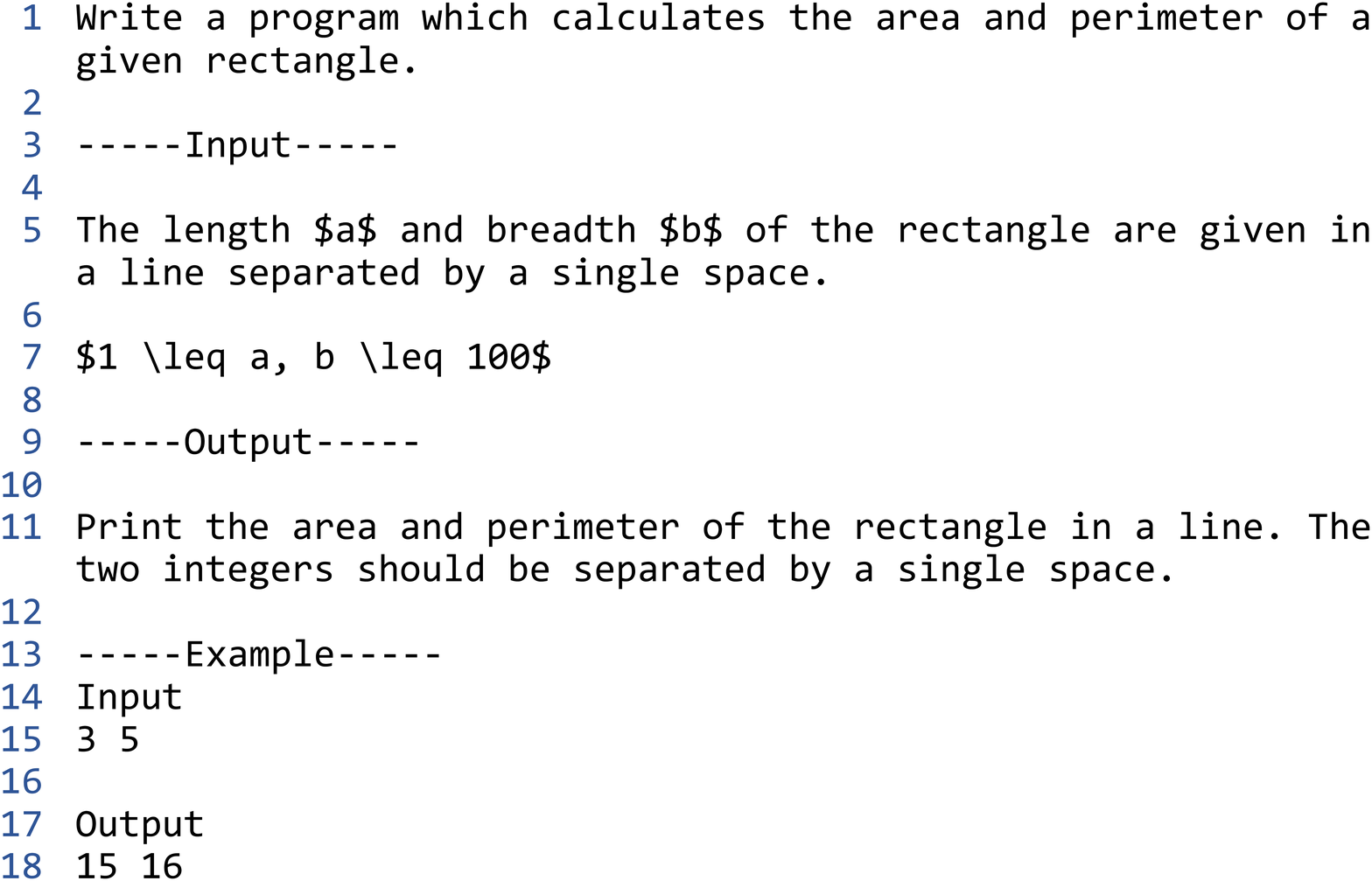}
    \caption{Example of APPS-inspired problem (ITP1\_1\_C).}
    \label{fig:apps-problem}
\end{figure*}

\begin{figure*}[ht]
    \centering
    \includegraphics[width=0.8\linewidth]{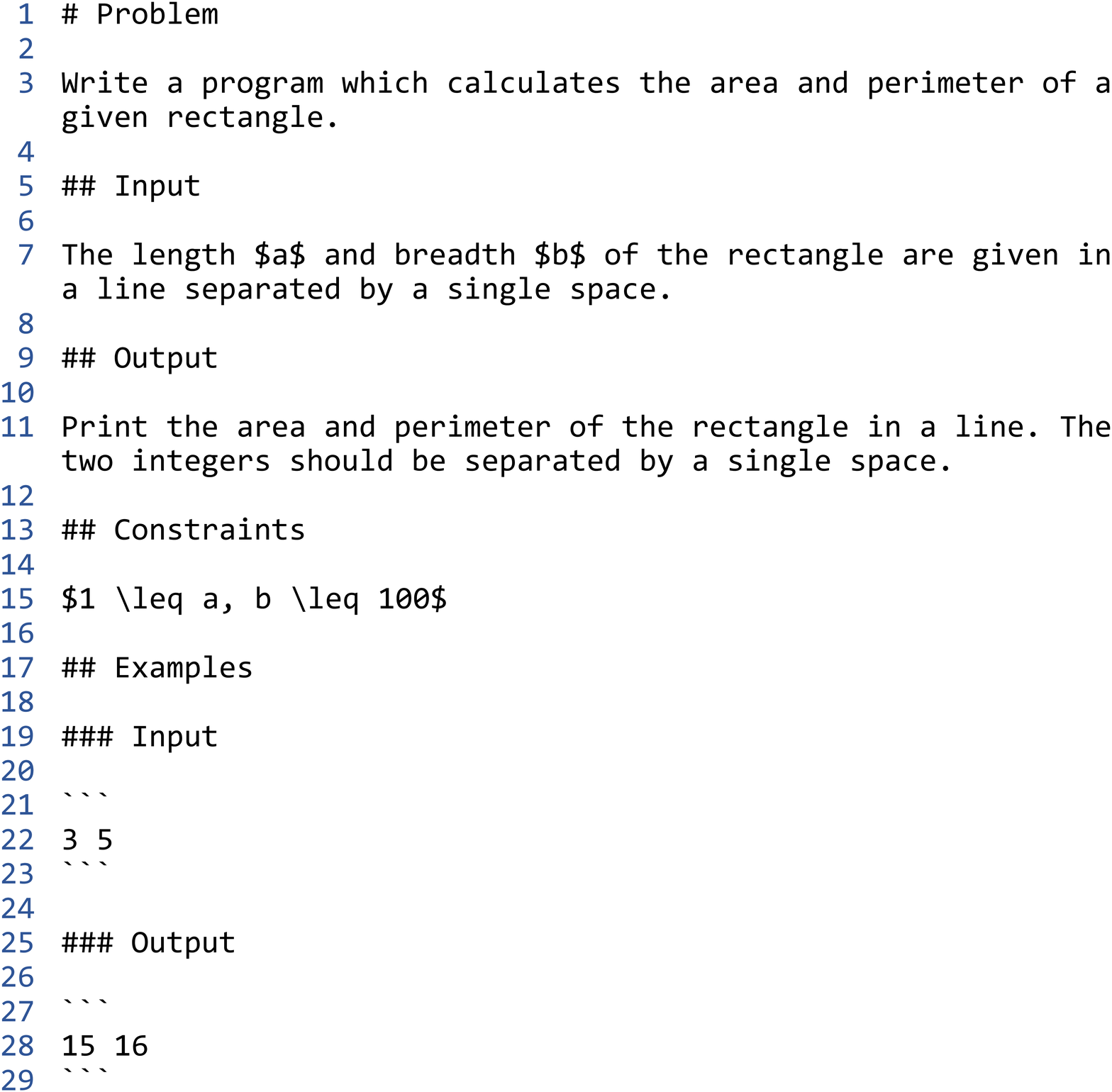}
    \caption{Example of fully Markdown-formatted problem (ITP1\_1\_C).}
    \label{fig:markdown-problem}
\end{figure*}

\begin{figure*}[ht]
    \centering
    \includegraphics[width=0.8\linewidth]{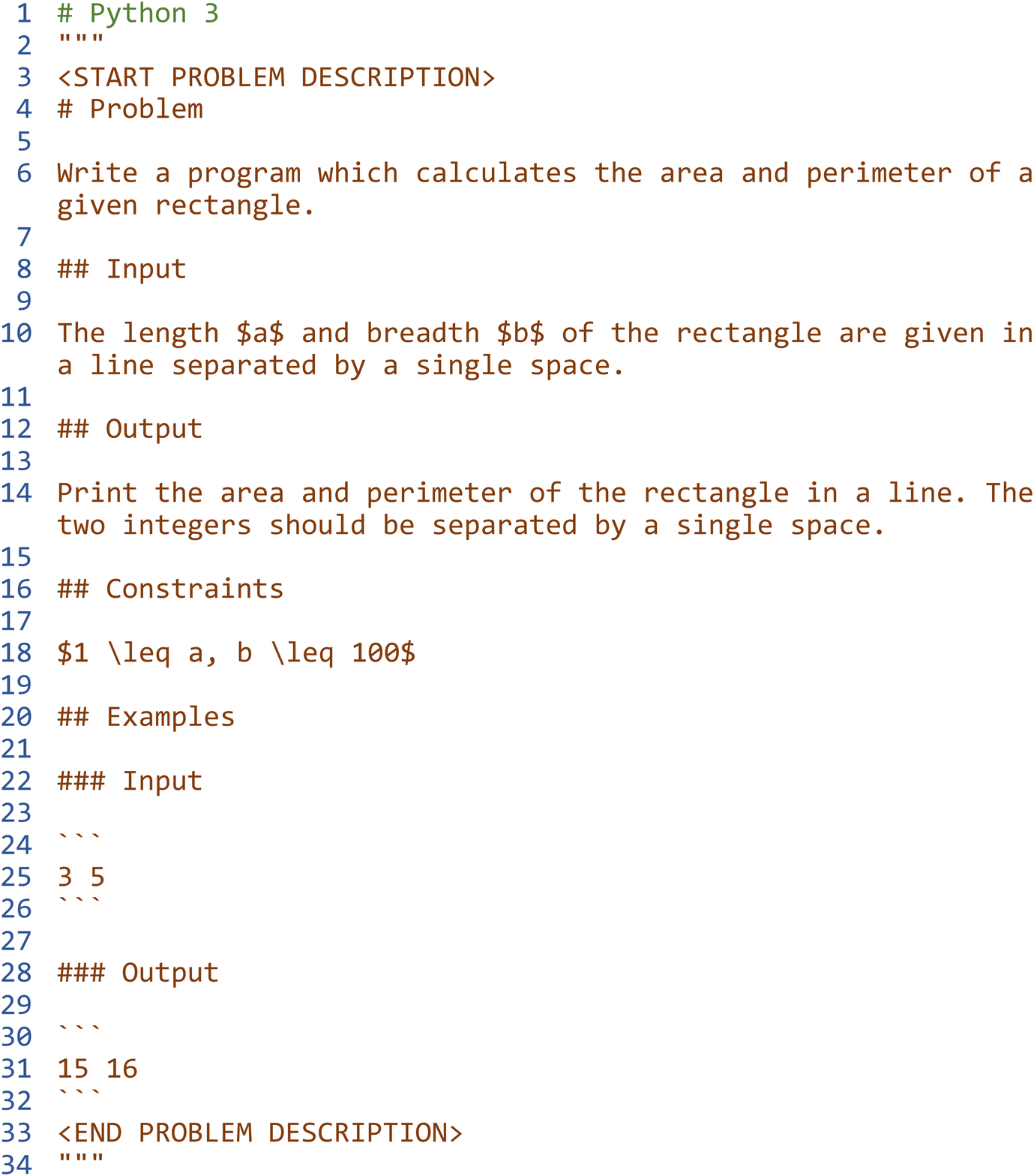}
    \caption{Example of formatted prompt of fully Markdown-formatted problem (ITP1\_1\_C).}
    \label{fig:prompt}
\end{figure*}

\end{document}